\documentclass[lettersize,journal]{IEEEtran}
\pdfoutput=1
\usepackage[T1]{fontenc}

\usepackage{amsmath,amsfonts}
\usepackage{algorithmic}
\usepackage{array}
\makeatletter
\let\MYcaption\@makecaption
\makeatother
\usepackage[font=footnotesize]{subcaption}
\makeatletter
\let\@makecaption\MYcaption
\makeatother

\usepackage{textcomp}
\usepackage{stfloats}
\usepackage{url}
\usepackage{verbatim}
\usepackage{graphicx}
\usepackage[style=ieee,
            doi=false,
            url=false,
            mincitenames=1,
            maxcitenames=1,
            natbib=true,
            minbibnames=6,
            maxbibnames=6,
            backend=biber]{biblatex}
\usepackage{csquotes} 
\AtBeginBibliography{\small}
\usepackage{balance}

\usepackage{glossaries}

\newglossaryentry{sensorgm}
{
  name=sensor measurement grid map,
  description={sensor measurement grid map}
}
\newglossaryentry{fusedgm}
{
  name=fused measurement grid map,
  description={fused measurement grid map}
}
\newglossaryentry{filteredgm}
{
  name=filtered grid map,
  description={filtered grid map}
}
\newglossaryentry{cam}
{
  name=camera,
  description={camera }
}
\newglossaryentry{monocam}
{
  name=monocular camera,
  description={monocular camera }
}
\newglossaryentry{stereocam}
{
  name=stereo camera,
  description={stereo camera }
}
\newglossaryentry{ground}
{
  name=ground semantics,
  description={occupancy dynamics}
}
\newglossaryentry{occdyn}
{
  name=occupancy dynamics,
  description={occupancy dynamics}
}
\newglossaryentry{occsem}
{
  name=occupancy semantics,
  description={occupancy semantics}
}
\newglossaryentry{semantickitti}
{
  name=SemanticKitti,
  description={SemanticKitti}
}

\usepackage[pdfstartview=XYZ,
bookmarks=true,
colorlinks=true,
linkcolor=black,
urlcolor=black,
citecolor=black,
bookmarks=true,
linktocpage=true, 
hyperindex=true
]{hyperref}
\usepackage{orcidlink}

\usepackage[capitalize]{cleveref}
\usepackage{cases}
\usepackage{rotating}
\usepackage{tabularx}
\usepackage{ragged2e}
\usepackage{forest}

\usepackage{longtable}
\usepackage{acro}

\DeclareAcronym{2D}{
    short = 2D,
    long = two-dimensional,
}
\DeclareAcronym{3D}{
    short = 3D,
    long = three-dimensional,
}
\DeclareAcronym{AD}{
    short = AD,
    long = automated driving,
}
\DeclareAcronym{ADAM}{
    short = ADAM,
    long = Adaptive Moment Estimation,
}
\DeclareAcronym{AI}{
    short = AI,
    long = artifical intelligence,
}
\DeclareAcronym{ANN}{
    short = ANN,
    long = artificial neural network,
}
\DeclareAcronym{AOE}{
    short = AOE,
    long = average orientation error,
}
\DeclareAcronym{AORE}{
    short = AORE,
    long = average odometry rotation error,
}
\DeclareAcronym{AOTE}{
    short = AOTE,
    long = average odometry translation error,
}
\DeclareAcronym{AP}{
    short = AP,
    long = average precision,
}
\DeclareAcronym{AR}{
    short = AR,
    long = average recall,
}
\DeclareAcronym{ARE}{
    short = ARE,
    long = average rotation error,
}
\DeclareAcronym{ASE}{
    short = ASE,
    long = average scale error,
}
\DeclareAcronym{ATE}{
    short = ATE,
    long = average translation error,
}
\DeclareAcronym{AVE}{
    short = AVE,
    long = average velocity error,
}
\DeclareAcronym{BBA}{
    short = BBA,
    long = basic belief assignment,
}
\DeclareAcronym{BCE}{
    short = BCE,
    long = binary cross entropy,
}
\DeclareAcronym{BiFPN}{
    short = BiFPN,
    long = Bidirectional Feature Pyramid Network,
}
\DeclareAcronym{BN}{
    short = BN,
    long = batch normalization,
}
\DeclareAcronym{CCL}{
    short = CCL,
    long = connected-components labeling,
}
\DeclareAcronym{CE}{
    short = CE,
    long = cross entropy,
}
\DeclareAcronym{CNN}{
    short = CNN,
    long = convolutional neural network,
}
\DeclareAcronym{DAG}{
    short = DAG,
    long = directed acyclic graph,
}
\DeclareAcronym{DBSCAN}{
    short = DBSCAN,
    long = density-based spatial clustering of applications with noise,
}
\DeclareAcronym{DEVN}{
    short = DEVN,
    long = directed evidential network with conditional belief functions,
}
\DeclareAcronym{DL}{
    short = DL,
    long = deep learning,
}
\DeclareAcronym{eIoU}{
    short = eIoU,
    long = evidential intersection over union,
}
\DeclareAcronym{ENC}{
    short = ENC,
    long = evidential network with conditional belief functions,
}
\DeclareAcronym{FLOP}{
    short = FLOP,
    long = floating-point operation,
}
\DeclareAcronym{FOD}{
    short = FOD,
    long = frame of discernment,
    long-plural-form = frames of discernment
}
\DeclareAcronym{FOV}{
    short = FOV,
    long = field of view,
}
\DeclareAcronym{FPN}{
    short = FPN,
    long = Feature Pyramid Network,
}
\DeclareAcronym{FPS}{
    short = FPS,
    long = frames per seconds,
}
\DeclareAcronym{GNC}{
    short = GNC,
    long = Graduated Non-Convexity,
}
\DeclareAcronym{GMC}{
    short = GMC,
    long = Geman McClure,
}
\DeclareAcronym{GPS}{
    short = GPS,
    long = Global Positioning System,
}
\DeclareAcronym{GPU}{
    short = GPU,
    long = Graphics Processing Unit,
}
\DeclareAcronym{IMU}{
    short = IMU,
    long = Inertial Measurement Unit,
}
\DeclareAcronym{IoU}{
    short = IoU,
    long = intersection over union,
}
\DeclareAcronym{KLD}{
    short = KLD,
    long = Kullback-Leiber divergence,
}
\DeclareAcronym{lidar}{
    short = LiDAR,
    long = light detection and ranging,
}
\DeclareAcronym{LLS}{
    short = LLS,
    long = linear Least-Squares,
}
\DeclareAcronym{LS}{
    short = LS,
    long = Least-Squares,
}
\DeclareAcronym{mAOE}{
    short = mAOE,
    long = mean average orientation error,
}
\DeclareAcronym{mAORE}{
    short = mAORE,
    long = mean average odometry rotation error,
}
\DeclareAcronym{mAOTE}{
    short = mAOTE,
    long = mean average odometry translation error,
}
\DeclareAcronym{mAP}{
    short = mAP,
    long = mean average precision,
}
\DeclareAcronym{mARE}{
    short = mARE,
    long = mean average rotation error,
}
\DeclareAcronym{mASE}{
    short = mASE,
    long = mean average scale error,
}
\DeclareAcronym{mATE}{
    short = mATE,
    long = mean average translation error,
}
\DeclareAcronym{mAVE}{
    short = mAVE,
    long = mean average velocity error,
}
\DeclareAcronym{MIB}{
    short = MIB,
    long = multi-instance Bernoulli,
}
\DeclareAcronym{ML}{
    short = ML,
    long = machine learning,
}
\DeclareAcronym{NMS}{
    short = NMS,
    long = non-maximum suppression,
}
\DeclareAcronym{OLS}{
    short = OLS,
    long = ordinary Least-Squares,
}
\DeclareAcronym{PCR}{
    short = PCR,
    long = partial conflict redistribution
}
\DeclareAcronym{PDF}{
    short = PDF,
    long = probability density function,
}
\DeclareAcronym{PHD}{
    short = PHD,
    long = probability hypothesis density,
}
\DeclareAcronym{PHD/MIB}{
    short = PHD/MIB,
    long = probability hypothesis density / multi-instance Bernoulli,
}
\DeclareAcronym{radar}{
    short = RaDAR,
    long = radio detection and ranging,
}
\DeclareAcronym{RFS}{
    short = RFS,
    long = random finite set,
}
\DeclareAcronym{RE}{
    short = RE,
    long = rotational error,
}
\DeclareAcronym{ReLU}{
    short = ReLU,
    long = rectified linear unit,
}
\DeclareAcronym{RPN}{
    short = RPN,
    long = region proposal network,
}
\DeclareAcronym{SGD}{
    short = SGD,
    long = stochastic gradient descent,
}
\DeclareAcronym{sonar}{
    short = Sonar,
    long = Sound Navigation and Ranging,
}
\DeclareAcronym{TE}{
    short = TE,
    long = translational error,
}
\DeclareAcronym{TLS}{
    short = TLS,
    long = truncated Least-Squares,
}
\DeclareAcronym{UBS}{
    short = UBS,
    long = uniform B-spline,
}
\DeclareAcronym{ER}{
    short = ER,
    long = evidential reasoning,
}
\DeclareAcronym{WLS}{
    short = WLS,
    long = weighted Least-Squares,
}

\usepackage{amsfonts, amsmath, amssymb, amsthm}
\usepackage{mathdots}
\usepackage{mathtools}

\usepackage{bm}

\theoremstyle{definition}
\newtheorem{definition}{Definition}

\newcommand{\func}[1]{\mathrm{#1}}

\DeclarePairedDelimiter\norm{\lVert}{\rVert}

\DeclarePairedDelimiterX{\inp}[2]{\langle}{\rangle}{#1, #2}

\newcommand{\leb}[1]{\mu(#1)}

\newcommand{\real}{\mathbb{R}}

\newcommand{\dx}[1]{\func{d}#1}
\newcommand{\bba}[1]{\func{m}(#1)}

\newcommand{\powerset}[1]{\mathcal{P}(#1)}

\newcommand{\perm}{\rho}
\newcommand{\frameg}{\Omega_g}

\newcommand{\cl}{\mathcal{S}}
\newcommand{\tf}{\mathcal{T}}
\newcommand{\rect}{\mathcal{R}}
\newcommand{\calib}{\mathcal{K}}
\newcommand{\measurement}{\mathcal{M}}

\newcommand{\grid}{\mathcal{G}}
\newcommand{\gridxy}{\mathcal{G}_{xy}}
\newcommand{\cell}{C}

\newcommand{\image}{\mathrm{f}}

\newcommand{\fbba}{\mathrm{m}}

\newcommand{\hist}{\mathrm{h}}
\newcommand{\gm}{\mathrm{g}}
\newcommand{\occ}{\mathrm{occ}}

\newcommand{\lettersem}{s}

\newcommand{\frameo}{\Omega_\lettersem}

\newcommand{\ovoid}{V_\lettersem}
\newcommand{\ofree}{F_\lettersem}
\newcommand{\ooccupied}{O}
\newcommand{\ocar}{O_{\mathrm{car}}}
\newcommand{\otwowheeler}{O_{\mathrm{tw}}}
\newcommand{\opedestrian}{O_{\mathrm{ped}}}
\newcommand{\omovable}{O_{\mathrm{om}}}
\newcommand{\ononmovable}{O_{\mathrm{im}}}

\newcommand{\range}{\mathrm{range}}
\newcommand{\normals}{\mathrm{normals}}
\newcommand{\height}{\mathrm{height}}

\newcommand{\distxy}{\mathrm{distXY}}
\newcommand{\sem}{\mathrm{sem}}

\newcommand{\conf}{\mathrm{conf}}

\newcommand{\pos}{\vec{p}}

\newcommand{\nvec}{\vec{n}}

\newcommand{\pr}[1]{\mathrm{Pr}(#1)}

\newcommand{\fpr}{\mathrm{Pr}}

\newcommand{\defeq}{\vcentcolon=}

\newcommand{\FP}{\mathrm{FP}}

\newcommand{\trans}[1]{\ensuremath{#1^{\scriptscriptstyle T}}}

\newcommand{\flatw}{\mathrm{flat}}
\newcommand{\spline}{\mathrm{spline}}

\makeatletter
\newcommand{\ocap}{\mathbin{\mathpalette\make@circled\smallcap}}
\newcommand{\make@circled}[2]{%
  \ooalign{$\m@th#1\smallbigcirc{#1}$\cr\hidewidth$\m@th#1#2$\hidewidth\cr}%
}
\newcommand{\smallbigcirc}[1]{%
  \vcenter{\hbox{\scalebox{0.77778}{$\m@th#1\bigcirc$}}}%
}
\newcommand{\smallcap}{%
  \vcenter{\hbox{\scalebox{0.6}{$\cap$}}}%
}
\makeatother

\usepackage{tikz}
\usepackage{ifthen}

\usepackage[inkscapepath=out/svg-inkscape/, notransparent]{svg}
\svgpath{figures/}

\graphicspath{{figures/},{tikz/}}

\usepackage{pgfplots}
\usepackage{stackengine}
\usepackage{tikzexternal}
\usepgfplotslibrary{
    colorbrewer,
    groupplots,
}

\pgfplotsset{
    bar cycle list/.style={
            cycle list name=Set1,
            every axis plot/.append style={fill, thin},
        },
    compat=1.15,
    colormap/Spectral,
    colormap={reverse Spectral}{
            indices of colormap={
                    \pgfplotscolormaplastindexof{Spectral},...,0 of Spectral}
        },
    cycle list/Set1,
    cycle list name=Set1,
    enlarge x limits=0.05,
    enlarge y limits=0.05,
    every axis/.append style={semithick},
    every axis plot/.append style={thick},
    grid style={
            opacity=0.5,
        },
    height=0.25\linewidth,
    scale only axis=true,
    scaled x ticks=false,
    scaled y ticks=false,
    title style={
            anchor=north east,
            at={(0.99, 0.95)},
            fill=white,
            fill opacity=0.5,
            text opacity=1,
        },
    width=0.9\linewidth,
}

\pgfkeys{/entities/.cd,
  file/.initial=,
  angle/.initial=0,
  width/.initial=,
  plot_ego/.initial=1,
  plot_hsv/.initial=1,
  roi_x_min/.initial=-50,
  roi_y_min/.initial=-50,
  roi_x_max/.initial=50,
  roi_y_max/.initial=50,
  hsv_xmin/.initial=40,
  hsv_xmax/.initial=46,
  hsv_ymin/.initial=-46,
  hsv_ymax/.initial=-40,
  }
\def\set@keys#1{
  \pgfkeys{/entities/.cd,#1}}
\def\get#1{
  \pgfkeysvalueof{/entities/#1}}
\newcommand{\quadgrid}[1]{
\begin{tikzpicture}
    \set@keys{#1}
    \begin{axis}[
            axis on top,
            major grid style={very thin},
            enlargelimits=false,
            grid=major,
            height=\get{width}\linewidth,
            width=\get{width}\linewidth,
            ticklabel style = {font=\tiny}
        ]
        \addplot[plot graphics/node/.append style={yscale=-1,anchor=north west}] graphics[xmin=\get{roi_x_min},xmax=\get{roi_x_max},ymin=\get{roi_y_min},ymax=\get{roi_y_max}, includegraphics={trim=0 0  0 0,clip,keepaspectratio}] {\get{file}};
        \addplot[rotate around={\get{angle}:(0,0)}] graphics[xmin=-2,xmax=3.2,ymin=-1.3,ymax=1.3] {bertha/car};
    \end{axis}
\end{tikzpicture}
}

\newcommand{\quadgridhsv}[1]{
\begin{tikzpicture}
    \set@keys{#1}
    \begin{axis}[
            axis on top,
            major grid style={very thin},
            enlargelimits=false,
            grid=major,
            width=\get{width}\linewidth,
            height=(\get{roi_y_max}-\get{roi_y_min})/(\get{roi_x_max}-\get{roi_x_min})*\get{width}\linewidth,
            axis equal,
            ticklabel style = {font=\tiny}
        ]
        \addplot[plot graphics/node/.append style={yscale=-1,anchor=north west}] graphics[xmin=\get{roi_x_min},xmax=\get{roi_x_max},ymin=\get{roi_y_min},ymax=\get{roi_y_max}, includegraphics={trim=0 0  0 0,clip,keepaspectratio}] {\get{file}};
        \ifthenelse{\equal{\get{plot_ego}}{1}}
           {\addplot[rotate around={\get{angle}:(0,0)}] graphics[xmin=-2,xmax=3.2,ymin=-1.3,ymax=1.3] {bertha/car};}{}
        \ifthenelse{\equal{\get{plot_hsv}}{1}}
            {\addplot graphics[xmin=\get{hsv_xmin},xmax=\get{hsv_xmax},ymin=\get{hsv_ymin},ymax=\get{hsv_ymax}] {hsv_wheel};}{}
    \end{axis}
\end{tikzpicture}
}

\pgfkeys{/wg/entities/.cd,
  file/.initial=,
  width/.initial=,
  }
\def\wgset@keys#1{
  \pgfkeys{/wg/entities/.cd,#1}}
\def\wgget#1{
  \pgfkeysvalueof{/wg/entities/#1}}
\newcommand{\widegrid}[1]{
\begin{tikzpicture}
    \wgset@keys{#1}
    \begin{axis}[
            axis on top,
            major grid style={very thin},
            enlargelimits=false,
            grid=major,
            height=0.4*\wgget{width}\linewidth,
            width=\wgget{width}\linewidth,
            xtick={-40, -20, ..., 40.0001},
            ytick={-40, -20, ..., 40.0001},
            xticklabels={-40, -20, ..., 40},
            yticklabels={-40, -20, ..., 40},
            ticklabel style = {font=\tiny}
        ]
        \addplot[plot graphics/node/.append style={yscale=-1,anchor=north west}] graphics[xmin=-50,xmax=50,ymin=-20,ymax=20, includegraphics={trim=0 300  0 300,clip,keepaspectratio}] {\wgget{file}};
        \addplot graphics[xmin=-2,xmax=3.2,ymin=-1.3,ymax=1.3] {bertha/car};
    \end{axis}
\end{tikzpicture}
}

\newcommand{\widegridcam}[1]{
\begin{tikzpicture}
    \wgset@keys{#1}
    \begin{axis}[
            axis on top,
            major grid style={very thin},
            enlargelimits=false,
            grid=major,
            height=0.67*\wgget{width}\linewidth,
            width=\wgget{width}\linewidth,
            xtick={-10, 0, ..., 40.0001},
            ytick={-40, -20, ..., 40.0001},
            xticklabels={-10, 0, ..., 40},
            yticklabels={-40, -20, ..., 40},
            ticklabel style = {font=\tiny}
        ]
        \addplot[plot graphics/node/.append style={yscale=-1,anchor=north west}] graphics[xmin=-10,xmax=50,ymin=-20,ymax=20, includegraphics={trim=400 300 0 300,clip,keepaspectratio}] {\wgget{file}};
        \addplot graphics[xmin=-2,xmax=3.2,ymin=-1.3,ymax=1.3] {bertha/car};
    \end{axis}
\end{tikzpicture}
}

\newcommand{\widegridfront}[1]{
\begin{tikzpicture}
    \wgset@keys{#1}
    \begin{axis}[
            axis on top,
            major grid style={very thin},
            enlargelimits=false,
            grid=major,
            height=0.5*\wgget{width}\linewidth,
            width=\wgget{width}\linewidth,
            xtick={-40, -20, ..., 40.0001},
            xticklabels={-40, -20, ..., 40},
            ytick={-15, 0, 15.0001},
            yticklabels={-15, 0, 15},
            ticklabel style = {font=\tiny}
        ]
        \addplot[plot graphics/node/.append style={yscale=-1,anchor=north west}] graphics[xmin=-10,xmax=50,ymin=-15,ymax=15, includegraphics={trim=400 350  0 350,clip,keepaspectratio}] {\wgget{file}};
        \addplot graphics[xmin=-2,xmax=3.2,ymin=-1.3,ymax=1.3] {bertha/car};
    \end{axis}
\end{tikzpicture}
}

\newcommand{\widegridur}[1]{
\begin{tikzpicture}
    \wgset@keys{#1}
    \begin{axis}[
            axis on top,
            major grid style={very thin},
            enlargelimits=false,
            grid=major,
            height=2.5*\wgget{width}\linewidth,
            width=\wgget{width}\linewidth,
            xtick={-40, -20, ..., 40.0001},
            ytick={-40, -20, ..., 40.0001},
            xticklabels={-40, -20, ..., 40},
            yticklabels={-40, -20, ..., 40},
            ticklabel style = {font=\tiny}
        ]
        \addplot[plot graphics/node/.append style={yscale=-1,anchor=north west}]  graphics[xmin=-20,xmax=20,ymin=-50,ymax=50, includegraphics={trim=0 300  0 300,clip,keepaspectratio,angle=-90,origin=c}] {\wgget{file}};
        \addplot[rotate around={90:(0,0)}] graphics[xmin=-2,xmax=3.2,ymin=-1.3,ymax=1.3] {bertha/car};
    \end{axis}
\end{tikzpicture}
}

\newcommand{\widegridhsv}[1]{
\begin{tikzpicture}
    \wgset@keys{#1}
    \begin{axis}[
            axis on top,
            major grid style={very thin},
            enlargelimits=false,
            grid=major,
            height=0.4*\wgget{width}\linewidth,
            width=\wgget{width}\linewidth,
            xtick={-40, -20, ..., 40.0001},
            ytick={-40, -20, ..., 40.0001},
            xticklabels={-40, -20, ..., 40},
            yticklabels={-40, -20, ..., 40},
            ticklabel style = {font=\tiny}
        ]
        \addplot[plot graphics/node/.append style={yscale=-1,anchor=north west}] graphics[xmin=-50,xmax=50,ymin=-20,ymax=20, includegraphics={trim=0 300  0 300,clip,keepaspectratio}] {\wgget{file}};
        \addplot graphics[xmin=-2,xmax=3.2,ymin=-1.3,ymax=1.3] {bertha/car};
        \addplot graphics[xmin=42,xmax=48,ymin=-18,ymax=-12] {hsv_wheel};
    \end{axis}
\end{tikzpicture}
}

\newcommand{\widegridhsvur}[1]{
\begin{tikzpicture}
    \wgset@keys{#1}
    \begin{axis}[
            axis on top,
            major grid style={very thin},
            enlargelimits=false,
            grid=major,
            height=2.5*\wgget{width}\linewidth,
            width=\wgget{width}\linewidth,
            xtick={-40, -20, ..., 40.0001},
            ytick={-40, -20, ..., 40.0001},
            xticklabels={-40, -20, ..., 40},
            yticklabels={-40, -20, ..., 40},
            ticklabel style = {font=\tiny}
        ]
        \addplot[plot graphics/node/.append style={yscale=-1,anchor=north west}] graphics[xmin=-20,xmax=20,ymin=-50,ymax=50, includegraphics={trim=0 300  0 300,clip,keepaspectratio,angle=-90,origin=c}] {\wgget{file}};
        \addplot[rotate around={90:(0,0)}] graphics[xmin=-2,xmax=3.2,ymin=-1.3,ymax=1.3] {bertha/car};
        \addplot[rotate around={90:(15,-45)}] graphics[xmin=12,xmax=18,ymin=-48,ymax=-42] {hsv_wheel};
    \end{axis}
\end{tikzpicture}
}

\pgfkeys{/map/entities/.cd,
  min/.initial=0,
  max/.initial=1,
  width/.initial=1,
  }
\def\mapset@keys#1{
  \pgfkeys{/map/entities/.cd,#1}}
\def\mapget#1{
  \pgfkeysvalueof{/map/entities/#1}}
\newcommand{\colorbar}[1]{
\begin{tikzpicture}
    \mapset@keys{#1}
    \pgfplotscolorbardrawstandalone[ 
        colormap={example}{
            rgb255(0cm)=(48,18,59)
            rgb255(1cm)=(70,107,227)
            rgb255(2cm)=(40,188,235)
            rgb255(3cm)=(50,242,152)
            rgb255(4cm)=(164,252,60)
            rgb255(5cm)=(238,207,58)
            rgb255(6cm)=(251,126,33)
            rgb255(7cm)=(208,47,5)
            rgb255(8cm)=(122,4,3)
        },
        colorbar horizontal,
        point meta min=\mapget{min},
        point meta max=\mapget{max},
        colorbar style={
            width=\mapget{width}\linewidth,
            height=5pt,
            draw=white
        }
    ]
\end{tikzpicture}
}

\definecolor{colorstreet}{RGB}{128, 64, 128}
\definecolor{colorsidewalk}{RGB}{219, 87, 213}
\definecolor{colorterrain}{RGB}{152, 251, 152}
\definecolor{colorcar}{RGB}{0, 0, 142}
\definecolor{colorcyclist}{RGB}{119, 11, 32}
\definecolor{colorped}{RGB}{220, 20, 60}
\definecolor{colormov}{RGB}{0, 80, 100}
\definecolor{colornonov}{RGB}{102, 102, 156}
\definecolor{colorocc}{RGB}{0, 0, 0}
\definecolor{colorfree}{RGB}{240, 240, 240}

\newcommand{\colorbarpx}{
    \begin{tikzpicture}
        \centering
        \coordinate (p);
        \foreach \name/\cl/\tcl in {
            Street/colorstreet/white,
            Sidewalk/colorsidewalk/white,
            Ground other/colorterrain/black,
            Car/colorcar/white,
            Immobile/colornonov/white}
        {
          \node[rectangle, text=\tcl, fill=\cl, draw=white, minimum height=1pt, minimum width=30pt, anchor=west]
          (n) at (p) {\name};
          \coordinate (p) at ([xshift=-1pt]n.east);
        }
    \end{tikzpicture}
}

\newcommand{\colorbarpxslim}{
    \begin{tikzpicture}
        \centering
        \coordinate (p);
        \foreach \name/\cl/\tcl in {
            Street/colorstreet/white,
            Sidewalk/colorsidewalk/white,
            Ground other/colorterrain/black,
            Car/colorcar/white,
            Immobile/colornonov/white}
        {
          \node[rectangle, text=\tcl, fill=\cl, draw=white, minimum height=1pt, minimum width=60pt, anchor=north]
          (n) at (p) {\name};
          \coordinate (p) at (n.south);
        }
    \end{tikzpicture}
}

\newcommand{\colorbarevslim}{
    \centering
    \begin{tikzpicture}
        \coordinate (p);
        \foreach \name/\cl/\tcl in {
            \vphantom{Op}Street/colorstreet/white,
            \vphantom{Op}Sidewalk/colorsidewalk/white,
            \vphantom{Op}Ground other/colorterrain/black,
            \vphantom{Op}Car/colorcar/white,
            \vphantom{Op}Two-wheeler/colorcyclist/white}
        {
          \node[rectangle, text=\tcl, fill=\cl, draw=white, minimum height=1pt, minimum width=48pt, anchor=west]
          (n) at (p) {\name};
          \coordinate (p) at ([xshift=-1pt]n.east);
        }
        \coordinate (p) at (0,-10pt);
        \foreach \name/\cl/\tcl in {
            \vphantom{Op}Pedestrian/colorped/white,
            \vphantom{Op}Other mobile/colormov/white,
            \vphantom{Op}Immobile/colornonov/white,
            \vphantom{Op}Occupied/colorocc/white,
            \vphantom{Op}Free/colorfree/black}
        {
          \node[rectangle, text=\tcl, fill=\cl, draw=white, minimum height=1pt, minimum width=48pt, anchor=west]
          (n) at (p) {\name};
          \coordinate (p) at ([xshift=-1pt]n.east);
        }
    \end{tikzpicture}
}

\newcommand{\colorbarevoccstack}{
    \fontsize{6pt}{9pt}\selectfont
    \coordinate (p) at (-3.22, 0);
    \node[rectangle, yshift=2.45cm, text=white, fill=colorcar, fill opacity=0.9, minimum height=10pt, minimum width=53pt, anchor=north]
    (n) at (p) {Car};
    \coordinate (p) at (n.south);
    \foreach \name/\cl/\tcl in {
        Two-wheeler/colorcyclist/white,
        Pedestrian/colorped/white,
        Other mobile/colormov/white,
        Immobile/colornonov/white,
        Occupied/colorocc/white,
        Free/colorfree/black}
    {
        \node[rectangle, text=\tcl, fill=\cl, fill opacity=0.9, minimum height=10pt, minimum width=53pt, anchor=north]
        (n) at (p) {\name};
        \coordinate (p) at (n.south);
    }
}

\newcommand{\colorbarevgroundstack}{
    \fontsize{6pt}{9pt}\selectfont
    \coordinate (p) at (-3.22, 0);
    \node[rectangle, yshift=-1.25cm, text=white, fill=colorstreet, fill opacity=0.9, minimum height=10pt, minimum width=53pt, anchor=north]
    (n) at (p) {Street};
    \coordinate (p) at (n.south);
    \foreach \name/\cl/\tcl in {
        Sidewalk/colorsidewalk/white,
        Ground other/colorterrain/black}
    {
        \node[rectangle, text=\tcl, fill=\cl, fill opacity=0.9, minimum height=10pt, minimum width=53pt, anchor=north]
        (n) at (p) {\name};
        \coordinate (p) at (n.south);
    }
}

\definecolor{colorobjbl}{RGB}{0, 255, 0}
\definecolor{colorobjnbl}{RGB}{255, 0, 0}
\definecolor{colorgroundbl}{RGB}{0, 0, 255}
\definecolor{colorgroundnbl}{RGB}{0, 0, 0}

\definecolor{colorblack}{RGB}{0, 0, 0}
\definecolor{colorgray}{RGB}{150, 150, 150}
\definecolor{colorred}{RGB}{205, 34, 34}
\definecolor{colorgreen}{RGB}{63, 205, 34}
\definecolor{colorblue}{RGB}{34, 63, 205}

\newcommand{\colorbarblockingconftv}{
    \centering
    \begin{tikzpicture}
        \coordinate (p);
        \foreach \name/\cl/\tcl in {
            {$\xi_{\mathrm{FP, i}}$}/colorblue/white,
            {$\xi_{\mathrm{FN, i}}$}/colorred/white,
            {$\xi_{\mathrm{TP, i}}+\xi_{\mathrm{TN, i}}$}/colorblack/white,
            {\vphantom{$\xi_{\mathrm{FP, i}}$}excluded}/colorgray/black}
        {
          \node[rectangle,text = \tcl,fill = \cl,draw=white, minimum height=1pt,minimum width=50pt,anchor=west]
          (n) at (p) {\name};
          \coordinate (p) at ([xshift=-1pt]n.east);
        }
    \end{tikzpicture}
}

\newcommand{\colorbarnormals}{
    \centering
    \begin{tikzpicture}
        \coordinate (p);
        \foreach \name/\cl/\tcl in {
            {$\trans{(1,0,0)}$}/colorblue/white,
            {$\trans{(0,1,0)}$}/colorgreen/white,
            {$\trans{(0,0,1)}$}/colorred/white}
        {
          \node[rectangle,text = \tcl,fill = \cl,draw=white, minimum height=10pt,minimum width=50pt,anchor=west, scale=0.8]
          (n) at (p) {\name};
          \coordinate (p) at ([xshift=-1pt]n.east);
        }
    \end{tikzpicture}
}
\usepackage{siunitx}

\robustify\bfseries

\DeclareSIUnit{\bits}{bits}
\newcommand{\polar}{Polar}
\newcommand{\cartesian}{Cartesian}
\newcommand{\SG}{Sensor Grid}
\newcommand{\sg}{sensor grid}
\newcommand{\MG}{Measurement Grid}
\newcommand{\mg}{measurement grid}
\newcommand{\CG}{Cartesian Grid}
\newcommand{\cg}{\cartesian{} grid}

\newcommand{\lidarkitti}{Velodyne HDL-64E}

\begin{document}

\title{Mapping LiDAR and Camera Measurements in a Dual Top-View Grid Representation Tailored for Automated Vehicles}

\author{Sven Richter\(^{1,\ast}\)\,\orcidlink{0000-0002-5429-4476}, Frank Bieder\(^{1,2}\), Sascha Wirges\(^{3}\) and
		Christoph Stiller\(^{1,2}\)\,\orcidlink{0000-0003-4165-2075}%
\thanks{\(^1\) Authors are with the Institute of Measurement and Control Systems, Karlsruhe Institute of Technology (KIT), Karlsruhe, Germany.%
{\tt\small \{sven.richter, frank.bieder, stiller\}@kit.edu}}
\thanks{\(^2\) Authors are with the Mobile Perception Systems Group, FZI Research Center for Information Technology, Karlsruhe, Germany.}%
\thanks{\(^3\) Author is with the Bosch Center for Artifical Intelligence, Renningen, Germany.
{\tt\small sascha.wirges@de.bosch.com}}
\thanks{\(^\ast\) Corresponding author.}
}

\maketitle

  \IEEEpubid{\begin{minipage}{\textwidth}~\\[12pt] \centering%
	This work has been submitted to the IEEE for possible publication. Copyright may be transferred without notice, after which this version may no longer be accessible.
  \end{minipage}}
  \IEEEpubidadjcol

\begin{abstract}
    We present a generic evidential grid mapping pipeline designed for imaging sensors such as \acs{lidar} and \glspl{cam}.
    Our grid-based evidential model contains semantic estimates for cell occupancy and ground separately.
    We specify the estimation steps for input data represented by point sets, but mainly focus on input data represented by images such as disparity maps or \acs{lidar} range images.
    Instead of relying on an external ground segmentation only, we deduce occupancy evidence by analyzing the surface orientation around measurements.
    We conduct experiments and evaluate the presented method using \acs{lidar} and \gls{stereocam} data recorded in real traffic scenarios.
    Our method estimates cell occupancy robustly and with a high level of detail while maximizing efficiency and minimizing the dependency to external processing modules.
\end{abstract}


\section{Introduction}

Traffic scene perception is one of the key tasks in the development of software components for automated transportation systems. 
Imaging sensors such as \acf{lidar} and \glspl{cam} play an important role in this context.
\acp{lidar} measure the distance to reflecting surfaces and thus enable the detection of obstacles and free space.
In contrast to \acp{lidar}, \glspl{cam} do not measure the distance to reflecting surfaces directly.
Instead, depth estimates must be provided based on the sensor readings by utilizing Computer Vision algorithms.
\Glspl{stereocam} enable the deduction of range estimates by finding corresponding pixels using epipolar geometry and calculated disparity values, see e.g. \cite{sgm,patchmatch}.
The pixel disparities are then used to infer \ac{3D} pixel coordinates.
In recent years, deep neural networks have also been trained to predict range information based on monocular \cite{qiao2021vip,yuan2022new} or stereo images \cite{ganet,cheng2020hierarchical}.
Evidential top-view grid maps enable modeling free space, traffic participants and information on their semantics in a common representation. 
It is desired to combine measurements from heterogeneous sensors to cope with erroneous measurements by resolving measurement conflicts.
One way to facilitate evidential data fusion is to estimate grid maps based on measurements from different sensor sources and combine them on a grid map level.

\begin{figure}[!t]
    \centering
    \fontsize{8pt}{8pt}\selectfont
    \begin{tikzpicture}
        \node at (0,0) {\includesvg[width=\linewidth]{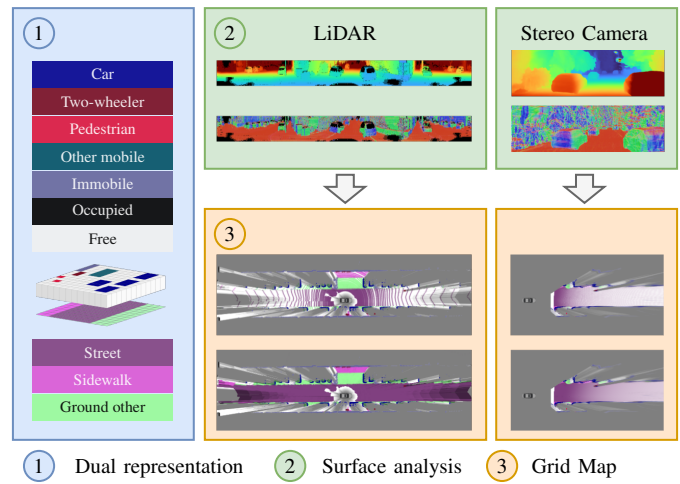}};
        \colorbarevoccstack
        \colorbarevgroundstack
    \end{tikzpicture}
    \caption{The proposed evidential grid mapping framework for input data given as images. 
            Range images from \ac{lidar} or disparity images from \glspl{stereocam} are transformed into a surface normal vector image. 
            This enables the deduction of occupancy evidence for detections on blocking surfaces. 
            The evidential grid map is estimated using two sensor models where semantic estimates for the ground and for occupancy may optionally be included.}
    \label{fig:overview}
\end{figure}
In this work, we propose a new method to estimate evidential top-view grid maps with measurements from imaging sensors, see \Cref{fig:overview}.
We present a novel dual representation that separates the environment into two layers in \Cref{sec:gm_model}.
The semantic state of the ground is estimated in the first layer and cell occupancy is modeled in the second layer.
Therefore, we propose two formal definitions of the term ``occupancy'' as a basis of our estimation.
\IEEEpubidadjcol
Then, we present a generic framework to estimate evidence masses for this evidential model in \Cref{sec:gm_framework}.
This framework is first specified for measurements represented as point sets and subsequently discussed in detail for measurements given as images.
By utilizing the organized structure of images, surface normal vectors can be calculated efficiently and occupancy evidence can be deduced for blocking surfaces.
For modelling spatial uncertainty, two sensor models are proposed that are calculated on sensor-dependent measurement grids such as \polar{} grids for \acp{lidar} and u/disparity grids for \glspl{stereocam}.
In \Cref{sec:experiments}, we present experiments with sensor measurements from a \ac{lidar} and a \gls{stereocam} in the publicly available \gls{semantickitti} dataset \cite{behley2019iccv}.
Besides showing qualitative results, we provide a detailed comparison study to demonstrate the robustness of our approach in challenging traffic scenarios.

\subsection{Evidence Theory}

In Evidence Theory~\cite{Shafer1976}, the \ac{BBA}
\begin{equation}
    \fbba \colon \powerset{\Omega} \rightarrow \left[ 0, 1 \right]~, \quad \bba{\emptyset} = 0, \quad \sum_{A \in \powerset{\Omega}} \bba{A} = 1
\end{equation}
assigns a degree of evidence to all possible combinations of hypotheses $\omega\subseteq\Omega$ contained in the powerset $\powerset{\Omega}$.
The hypotheses set $\Omega$ is called \ac{FOD} and consists of mutually excluding hypotheses of interest.

\subsection{Grid Maps}

The \ac{2D} grid $\grid = \mathcal{P}_1\times \mathcal{P}_2$ on the rectangular region of interest $\rect = I_1\times I_2 \subset \real^2$, where
\begin{align*}
    \mathcal{P}_i & = \{I_{i, k},\, k\in \{0,\dots,s_i-1\}\},                      \\
    I_{i, k}      & =[o_i + k\,\delta_i, o_i + (k+1)\,\delta_i),\quad i\in\{1, 2\}
\end{align*}
forms a partition of the interval $I_i$ with equidistant length $\delta_i \in\real$, origin $o_i\in \real$ and size $s_i\in\mathbb{N}$.
Hence, each grid cell $\cell\in\grid$ is a rectangle with side lengths $(\delta_1, \delta_2)$.

Grid mapping is the task of estimating a state $x$ in the state space $X$ for each grid cell $\cell\in\grid$.
Therefore, a grid map $\gm$ is a mapping
\begin{equation}
    \gm \colon \grid \rightarrow X
\end{equation}
assigning to each grid cell $\cell\in\grid$ an element in the state space $x\in X$.
The state space $X$ may encode any formalizable information on the local environment.

\subsection{Inverse Sensor Models}

When modeling a sensor measurement $m$ in a grid cell $\cell\in\grid$, spatial uncertainty is modeled by the inverse sensor model given by the conditional probability 
\begin{equation}
    \pr{x\in\cell\vert x_m},
\end{equation}
where $x_m$ is the measured position projected to the top-view grid and $x$ is the random variable representing the real position. 
In the remainder of this thesis, we write $\pr{\cell\vert m}$ for short. 
While the final representation is usually defined on a \cg{}, the inverse sensor model may be calculated in different coordinate systems such as \polar{}, u/distance or u/disparity grids.
The choice of the coordinate system depends on the sensor modalities and is made so that the grid can be aligned with the sensor measurement rays.

\section{Related Work}

\subsection{Grid Mapping with Range Sensors}

\citet{Elfes1989} proposed to model a range measurement recorded by a \acf{sonar} sensor by a \ac{2D}  Gaussian inverse sensor model where the two dimensions correspond to range and angle.
Based on the inverse sensor model he derived an occupancy profile that is recursively fused in a Bayesian framework.
\citet{Yguel2008} applied a simplified one-dimensional inverse sensor model to \ac{lidar} range measurements in a \polar{} grid.
They further focus on formulating the problem of switching coordinates from \polar{} to \cartesian{} mathematically and propose a suitable approximation that can be efficiently implemented on the \ac{GPU}.
\citet{Yu2014} handle conflicting measurements by first collecting reflections above a given height threshold in a \polar{} grid and subsequently transform the measurement counts into a \ac{BBA}.
They treat ground detections as sources of evidence for the hypothesis \emph{free} and apply backward extrapolation to deduce free space evidence to neighboring grid cells.
\citet{Porebski2020} presented a customizable inverse sensor model to calculate occupancy grid maps.
In order to be able to compute accurate probabilities in each grid cell, they proposed a cell selection process and apply either a Gaussian or an exponential distribution to compute the inverse sensor model.
Recently, \citet{vKempen2021} proposed an evidential occupancy grid mapping framework using end-to-end learning.
They generate synthetic \ac{lidar} point clouds based on simulated scenarios and train a deep neural network that is able to generate \ac{BBA} layers for the hypotheses \emph{occupied} and \emph{free} successfully modeling uncertainty.

\subsection{Grid Mapping with Cameras}

\citet{Badino2007} computed an occupancy grid map based on stereo disparity images and used the resulting grid representation to infer free space areas.
They compared Gaussian inverse sensor models on a \cartesian{}, \polar{} and u/disparity grid, respectively.
\citet{Yu2015} modeled free space in a v/disparity grid and occupancy separately in a u/disparity grid based on stereo measurements and subsequently combine both in an evidential occupancy grid map.
\citet{Valente2018} utilized a ground segmentation in a v/disparity grid and project obstacles in a u/disparity grid.
They apply a \ac{2D}  Gaussian inverse sensor model explicitly modeling errors in the stereo matching.
The vast amount of semantic segmentation frameworks in Computer Vision suggests including semantic estimates in Vision-based grid maps.
\citet{Giovani2015} added an occupancy refinement value denoting the semantic state as meta information to their grid map representation.
\citet{Thomas2019} incorporated semantic hypotheses in an evidential framework in order to estimate a geometric road model.
Their inverse sensor model considers confidences of the pixelwise semantic segmentation model and pixel location probabilities.

None of the above-mentioned publications models occupancy and semantic estimates in a joint evidential context.
In \cite{Richter2019,Richter2020,Richter2021} a sensor grid mapping pipeline was presented estimating a \ac{BBA} on a \ac{FOD} containing ground and object hypotheses for range sensors and \glspl{cam}.
Here, we present an advancement of the evidential model and rethink the \ac{BBA} estimation.
\section{The Evidential Grid Map Model} \label{sec:gm_model}

We define the driving corridor between the ground and the maximal height above ground $d_{z,\max}$.
This excludes high obstacles such as bridges and tree branches that do not interfere with the automated vehicle.
Let the surface of a traffic scene be implicitly given as $f_S(x, y, z) = 0$ where $x$, $y$ and $z$ are Cartesian coordinates in the vehicle coordinate system.
\begin{definition}\label{def:occupancy_1}
    Let $f_G\colon \real^2 \rightarrow \real$ be a function describing an approximation of the ground height and
    \begin{equation}
        D_\cell = \left\{z-f_G(x, y)\,\vert\, (x,y)\in\cell\colon f_S(x,y,z) = 0\right\}
    \end{equation}
    be the set containing all distances between ground surface and traffic scene surface in grid cell $\cell\in\grid$.
    The grid cell is called \emph{occupied}, if the traffic scene surface intersects with the driving corridor, i.e.
    \begin{equation}
        \sup\left(D_\cell\right) > 0,\quad \inf\left(D_\cell\right) < d_{z,\max},\label{eq:occupancy_1}
    \end{equation}
    where the supremum $\sup(A)$ is the smallest upper bound and the infimum $\inf(A)$ is the largest lower bound of an ordered set $A$.
    If both conditions in \Cref{eq:occupancy_1} are not fulfilled the cell is called \emph{free}.
\end{definition}
For the condition ``$\sup\left(D_\cell\right) > 0$'', the ground surface estimation must be very accurate to exclude all the measurements reflected on the ground.
Hence, a small tolerance margin $\delta_G > 0$ is added in some publications such as \cite{Wirges2021} so that the condition becomes ``$\sup\left(D_\cell\right) > \delta_G$''.

The second definition of the term occupancy is based on the unit normal vector of the surface $f_S(x, y, z) = 0$.
It can be shown that this normal vector $n$ is given by the gradient of $f_S$:
\begin{equation}
    \trans{(n_1, n_2, n_3)} = \frac{\nabla f_S}{|\nabla f_S|}.
\end{equation}
\begin{definition}\label{def:occupancy_2}
    A grid cell $\cell\in\grid$ is called
    \begin{itemize}
        \item \emph{occupied}, if the angle difference between the surface normal and the z-axis in the vehicle coordinate system exceeds a given threshold $b_T$, i.e.
        \begin{equation}
            \arccos(n_3) > b_T,
        \end{equation}
        and the traffic scene surface corresponding to the grid cell is not fully above the driving corridor, i.e. $\inf\left(D_\cell\right) < d_{z,\max}$.
        \item \emph{free}, if the ego vehicle can enter the underlying area in space, i.e. the complete driving corridor is free, or
        \item \emph{void}, if it is neither free nor occupied.
    \end{itemize}
\end{definition}
The cell state \emph{void} is attained in grid cells covered by obstacles where the surface is non-blocking.

In this work, we consider the semantic occupancy hypotheses ``car'', ``two-wheeler'', ``pedestrian'', ``other mobile obstacles'' or ``immobile obstacles'' and the ground hypotheses ``street'', ``sidewalk'' and ``other ground''.
Note that the occupancy hypotheses, e.g. occupied by a car, and the ground hypotheses, e.g. street, are not contradicting.
Therefore, we model occupancy and ground state in separate \acp{FOD}.

Occupancy is estimated within a cuboid whose footprint is given by the grid cell and the height interval $[0, d_{z,\max}]$ is defined by the driving corridor.
In particular, the occupancy \ac{FOD}
\begin{equation}
    \frameo \defeq \{ c, cy, p, m, nm, f, v \}
    \label{eq:fod_o}
\end{equation}
consists of the hypotheses listed in \Cref{tab:hypotheses_obstacle}.
\begin{figure}[!t]
    \centering
    \fontsize{8pt}{8pt}\selectfont
    \includesvg[width=.98\linewidth]{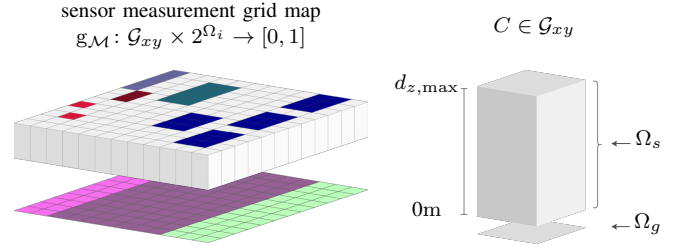}
    \caption{The evidential grid map model. 
            The dual grid map representation $\gm_\measurement$ contains the \ac{BBA} on the occupancy \ac{FOD} $\frameo$ and the ground \ac{FOD} $\frameg$.
            Occupancy is estimated in the driving corridor $[0, d_{z,\max}]$ above ground.}
    \label{fig:semantic_fod}
\end{figure}
\begin{table}[ht]
	\caption{The occupancy \ac{FOD}.}\label{tab:hypotheses_obstacle}
    \centering
	\begin{tabular}{lcc}
		Semantic class 	& Set & Letter \\
		\hline
		Occupied by\ldots      			    &       	&                   \\
		\ldots car 				            & $\{c\}$	&   $\ocar$         \\
		\ldots two-wheeler 			        & $\{cy\}$	&   $\otwowheeler$     \\
		\ldots pedestrian 		            & $\{p\}$   &   $\opedestrian$  \\
		\ldots other mobile object 	        & $\{om\}$  &   $\omovable$     \\
		\ldots immobile object              & $\{im\}$  &   $\ononmovable$  \\
        \ldots unknown object type          & $\{c, cy, p, om, im\}$  &   $\ooccupied$  \\
		\hline
		Free 			                    & $\{f\}$   &   $\ofree$    \\
		Neither occupied nor free, 	        & $\{v\}$   &   $\ovoid$       \\
	\end{tabular}
\end{table}
Note that this \ac{FOD} can be partitioned as $\frameo\setminus \ovoid = \ooccupied\,\dot\cup\,\ofree$ which is the separation used in classical occupancy grid mapping.

The ground \ac{FOD}
\begin{equation}
    \frameg \defeq \{ s, sw, t \}
    \label{eq:fod_g}
\end{equation}
describes the semantic state of the ground and consists of the hypotheses listed in \Cref{tab:hypotheses_ground}.
\begin{table}[ht]
	\caption{The ground \ac{FOD}.}\label{tab:hypotheses_ground}
    \centering
	\begin{tabular}{lc}
		Semantic class 	        &   Set     \\
		\hline
		Street 				    & $\{s\}$   \\
		Sidewalk 			    & $\{sw\}$  \\
		Other ground 		    & $\{t\}$
	\end{tabular}
\end{table}

The \ac{BBA} $\mathrm{m}$ on $\powerset{\Omega_i}, i\in\{\lettersem, g\}$ is then represented by the multi-layer grid map
\begin{equation}
    \gm_i \colon \gridxy \times \powerset{\Omega_i}\rightarrow [0, 1].
\end{equation}
The dual evidential representation is sketched in \Cref{fig:semantic_fod}.
It is based on the assumption that in traffic scenes each obstacle is placed on top of ground implying that every combination of hypotheses $\theta_1 \in \frameo$ and $\theta_2 \in \frameg$ are non contradicting and can thus happen simultaneously.
Note that the real world might be composed of several overlapping ground layers as e.g. at freeway exit ramps or bridges.
However, even in those scenarios the region of interest can be limited to the ground layer as there is no direct interaction between traffic participants on different ground layers.
The simplification of considering one occupancy layer only is justified by the fact that two objects placed on top of each other may be considered as one entity in the navigation module.
Therefore, the person sitting on a bicycle is considered as the one entity represented by the hypotheses ``occupied by a two-wheeler''.

\section{The Grid Mapping Framework} \label{sec:gm_framework}

\begin{figure}[!t]
    \centering
    \fontsize{8pt}{8pt}\selectfont
    \includesvg[width=\linewidth]{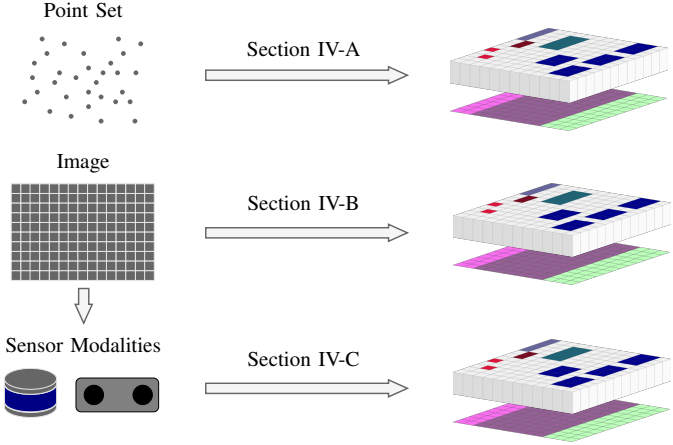}
    \caption{The methodology described in this section.
            The evidential grid map estimation is presented for point sets and images.
            The latter is applied to specific sensors considering their modalities. }
    \label{fig:sensor_gm_outline}
\end{figure}

Let $\measurement$ be a sensor measurement consisting of individual measurement elements $m\in\measurement$ with attached semantic label $\omega_m\in\powerset{\ooccupied}\cup\powerset{\frameg}$.
We present a generic framework for transforming the sensor measurement $\measurement$ to an evidential grid map $\gm_\measurement$.
The outline of the methodology described in this section is sketched in \Cref{fig:sensor_gm_outline}.

\subsubsection{Occupancy estimation}

Each measurement element $m$ provides an evidence for occupancy $\omega\subseteq\ooccupied$ or ground $\omega\subset\frameg$ depending on the attached semantic label $\omega_m$.
The evidential grid map for hypothesis $\omega$ in the \cg{} cell $\cell\in\gridxy$ is modeled as
\begin{equation}
    \gm_\measurement(\cell,\omega) = 1 - \prod\limits_{m\in\measurement} \pr{m\nrightarrow\omega,\cell}, \label{eq:bba}
\end{equation}
where $\pr{m\nrightarrow\omega,\cell}$ is the probability that the measurement element $m$ is not relevant for hypothesis $\omega$ in grid cell $\cell$.
It is calculated by considering the following four nested binary queries:
\begin{enumerate}
    \item Is the measurement element $m$ a true positive? 
        The false positive rate $p_\FP$ is a sensor dependent design parameter quantifying the likelihood of obtaining ghost detections.
    \item Was the measurement element $m$ recorded on an occupying surface? 
        Methods to calculate the according occupancy probability $p_\occ$ are discussed in \Cref{sec:sensor_gm_unstruct,sec:sensor_gm_struct}.
    \item Does the semantic label $\omega_m$ assigned to the measurement element $m$ match with the considered hypothesis $\omega$? 
        The according probability $p_{\omega}$ may be obtained from the confidences provided by the pixelwise semantic labeling algorithm.
        If no such information is available, $p_{\omega}$ may be set to one if $\omega_m = \omega$ and zero otherwise.
    \item Does the measurement element $m$ provide any evidence for grid cell $\cell$ based on its spatial uncertainty? 
        The according probability is given by the inverse sensor model $\pr{\cell\vert m}$, presented in \Cref{sec:sensor_gm_sensors}.
\end{enumerate}
Consequently, the calculation reads
\begin{align}
    \pr{m\nrightarrow\omega,\cell}   =  & p_\FP \nonumber \\                                      
                                        & + \left(1 - p_\FP\right)\cdot \left(1 - p_\occ\right) \nonumber \\
                                        & + \left(1 - p_\FP\right)\cdot p_\occ\cdot \left(1 - p_\omega\right) \nonumber \\
                                        & + \left(1 - p_\FP\right)\cdot p_\occ\cdot p_\omega\cdot \left(1 - \pr{\cell\vert m}\right).
                                        \label{eq:not_rel_obj_cell}
\end{align}
For \gls{ground} $\omega\subseteq\frameg$ the calculations are done analogously where the second query is negated as ground surfaces are assumed to be non-occupying.

\subsubsection{Free space estimation}

A grid cell is free, if no obstacles are present in a defined free space corridor.
The free space corridor is limited by the values $f_{z, \min},f_{z, \max}\in\real$ denoting the distance to the ground, where $0 \leq f_{z, \min} < f_{z, \max} \leq d_{z, \max}$.
This relation ensures that the free space corridor is part of the driving corridor.
The reason for defining another corridor specifically for the free space estimation is to allow traversing measurement parts of the driving corridor without providing free space evidence.
For instance, rays might traverse grid cells below cars.
The free space corridor is the height interval where it is very unlikely to have traversing measurements, if the cell is not free.
Evidence for the absence of obstacles is provided by measurement rays traversing the grid cell.
The free space evidence deduced from each traversing measurement ray is quantified as the ray height relative to the height of the free space corridor.
The ray permeability
\begin{equation}
    \perm = \frac{d_z}{f_{z,\max} - f_{z, \min}}. \label{eq:permeability}
\end{equation}
is then calculated as the ratio between the height portion $d_z$ covered by traversing measurement rays and the overall height of the free space corridor.
Furthermore, evidence for a cell not being free is obtained by any measurement that provides occupancy evidence. 
The \ac{BBA}
\begin{equation}
    \bba{\ofree} = \perm \cdot \left(1 - \sum\limits_{\psi\neq \ofree} m(\psi)\right) \label{eq:bba_free}
\end{equation}
for free space $\ofree$ is then calculated as the product of the ray permeability $\perm$ and the part of the \ac{BBA} mass that has not been assigned to any of the occupancy hypotheses $\omega\subseteq\ooccupied$.

\subsection{Grid Mapping with Point Sets} \label{sec:sensor_gm_unstruct}

Let the sensor measurement $\measurement$ be a point set, i.e. a measurement element $m\in\measurement$ is a detection coordinate indicating the presence of a reflecting surface with attached semantic label $\omega_m$.
Note that other information such as \ac{lidar} intensities or \ac{radar} Doppler measurements are omitted here as they are not considered in \Cref{eq:not_rel_obj_cell}.
Point set measurements may be obtained from range sensors such as \acp{radar}.
In point sets, no spatial relation between the elements is given.
The calculation of surface normal vectors used in \Cref{def:occupancy_2} requires finding neighboring elements which is computational expensive.
Therefore, occupancy is modeled according to \Cref{def:occupancy_1} which means that the detection point set is segmented into obstacle and ground detections based on a ground surface model such as presented in \cite{Wirges2021}.
In \Cref{eq:not_rel_obj_cell}, the occupancy probability $p_\occ$ is then set to one if $m$ was classified as occupying and to zero otherwise.

The ray permeability $\perm$ used for the \ac{BBA} estimation for the hypothesis \emph{free} $\ofree$ is approximated as 
\begin{equation}
    \perm \approx \frac{h_{\max} - h_{\min}}{f_{z,\max} - f_{z, \min}}, \label{eq:ray_perm_approx}
\end{equation}
where $h_{\min}, h_{\max} \in \real$ are the minimal and maximal measured heights of traversing measurement rays within the driving corridor.
Note that this approximation may differ from the real ray permeability significantly around obstacles not connected to the ground.

This grid mapping framework for point sets has the disadvantage that a ground surface estimation or an external ground segmentation module is required to estimate the occupancy probability $p_\occ$.
This occupies additional computational resources and may introduce errors.

\subsection{Grid Mapping with Images} \label{sec:sensor_gm_struct}

In this section, the generic grid mapping pipeline is put into concrete terms for measurements given as images.
The measurement images may be provided by different sensor types in different forms such as range images from \acp{lidar} or depth/disparity images from Cameras.

Throughout the processing steps, the representation is transformed to grids defined in different coordinate systems.
In particular, the following grids are considered:
\paragraph{\SG{}} 
The \sg{} $\grid_{uv}$ represents the measurement pattern of the sensor.
One sensor reading on $\grid_{uv}$
\begin{align*}
    \measurement = \{ & \image_\range \colon \grid_{uv} \rightarrow \real \cup \{\text{unknown}\},  \\
                    & \image_\sem   \colon \grid_{uv} \rightarrow \cl \cup \{\text{unknown}\} \},
\end{align*}
consists of a range measurement given by the mapping $\image_\range$ and potentially semantic estimates given by the mapping $\image_\sem$.
Here, $\cl = \ooccupied \cup \frameg$ is the set containing all singleton semantic hypotheses.
One sensor element $m\in\measurement$ is identified by the 3-tuple $(\cell, r_m, \omega_m)$ consisting of the \sg{} cell $\cell\in\grid{uv}$, the range measurement $r_m\in\real_{>0}$ and the semantic measurement $\omega_m\in\cl$.
The meaning of the range measurement $r_m$ depends on the sensor and may be the measured distance for \ac{lidar} sensors or the pixel disparity for \gls{stereocam}s.
Note that the sensor reading $\measurement$ marks the entry point of the estimation pipeline presented in this work and that there might be preprocessing steps required to obtain that information from the raw sensor measurements such as disparity calculation or pixelwise semantic labeling.
\paragraph{\MG{}} 
The \mg{} $\grid_{ur}$ consists of the horizontal \sg{} index $u$ in the first dimension and discretizes the range measurements interval of interest in the second dimension.
When collecting measurements from the \sg{} in the \mg{}, an orthographic projection along the upright \cartesian{} coordinate axis in the sensor coordinate system is performed.
An example for a \mg{} is a grid in \polar{} coordinates.
\paragraph{\CG{}} 
The final evidential grid map is defined on a \cartesian{} top-view grid $\gridxy$. 
It is defined on a sensor independent coordinate system where the origin is defined with respect to a fixed location on the ego vehicle such as the center of the rear axis.

To indicate the corresponding coordinate system, the~considered region of interest is subscripted analogously as $\rect_{uv}$, $\rect_{ur}$ and $\rect_{xy}$, respectively.
Furthermore, the~mappings
\begin{align*}
    \tf_{uv}^{ur} & \colon \rect_{uv}\rightarrow\rect_{ur},\quad \tf_{ur}^{xy} \colon \rect_{ur}\rightarrow\rect_{xy}
\end{align*}
are introduced for transforming coordinates from one system to~another.

\begin{figure}[!t]
    \centering
    \fontsize{8pt}{8pt}\selectfont
    \includesvg[width=\linewidth]{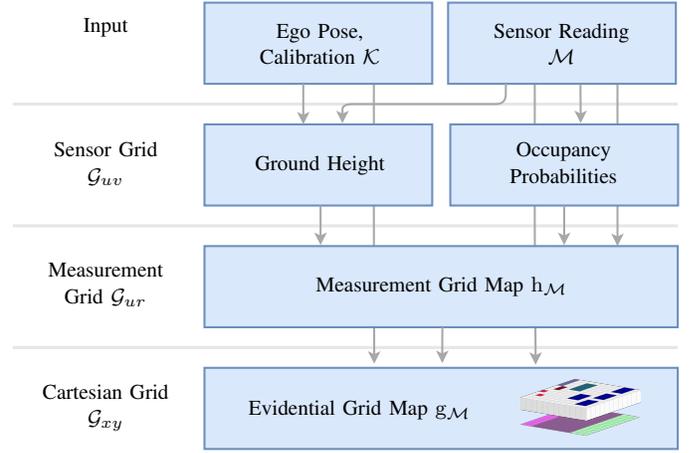}
    \caption{The processing blocks for calculating the evidential grid map based on input images.}
    \label{fig:sensor_gm_components}
\end{figure}

The individual processing blocks for calculating the evidential grid map $\gm_\measurement$ are depicted in \Cref{fig:sensor_gm_components}.
The extrinsic and intrinsic sensor calibrations as well as the 6-dimensional ego pose consisting of the \ac{3D} position and orientation are assumed to be known.
The first processing layer contains image processing steps on the \sg{} $\grid_{uv}$.
First, the surface normal vector is calculated for each measurement element based on the sensor calibration and the input range image.
Given the surface normal vectors, the occupancy probability $p_\occ$ from \Cref{eq:not_rel_obj_cell} can be computed.
Subsequently, each pixel is assigned a height above ground by propagating the heights of ground detections along each image column.
This information is later used for the free space estimation based on the vertical sensor ray coverage.
In the second processing layer a change of coordinates is applied from the \sg{} $\grid_{uv}$ to the top-view \mg{} $\grid_{ur}$.
This coordinate system is chosen according to the sensor characteristics, so that noise can be handled reasonably, and individual rays can be traced efficiently.
Here, individual sensor reflections are mapped into the \mg{} map $\hist_\measurement$ where evidence for occupancy and free space is accumulated.
Finally, the evidential grid map $\gm_\measurement$ is calculated based on the \mg{} map $\hist_\measurement$ in a common \cg{} $\gridxy$.
In the following, we explain the calculation steps on the three grids in detail.

\subsubsection{\SG{}} \label{subsec:gm_sensor_grid}

The surface at the reflection locations is analyzed to identify measurement elements that stem from occupying surfaces according to \Cref{def:occupancy_2}.
The decision if a measurement element stems from an occupying surface is made based on the surface normal vector at that location.
Instead of providing a binary classification of each measurement element into occupying and non-occupying, we calculate the probability $p_\occ\in[0, 1]$ that the surface reflecting the measurement element is occupying.
This occupancy probability $p_\occ$ can scale the resulting \ac{BBA} whereas a binary classification results in a loss of information.

Following \cite{KinectFusion}, a bilateral filter is applied to the measurement elements to eliminate noise while preserving edges.
Here, the geometric context plays an important role.
Averaging applied in the bilateral filter provides desired results if the measurement elements used for calculating the average have similar sources with zero-mean disturbances.
In order to increase the likelihood for this, the range image $\image_\range$ is transformed to images $\image_\height$ and $\image_\distxy$ representing the height of the measurement element relative to the sensor origin and the distance to the sensor origin projected to the XY-plane.
This selection is based on the assumption that environments in traffic scenes can be separated into sub-planes that are mostly oriented along the XY-plane or perpendicular to it.
This holds for the ground surface as well as many objects as buildings and traffic participants.
The bilateral filter is then applied to $\image_\height$ and $\image_\distxy$ with parameters $\sigma_{r, \height}$ and $\sigma_{r, \distxy}$ denoting the range standard deviation and $\sigma_{d, \height}$ and $\sigma_{d, \distxy}$ denoting the spatial standard deviation.

Based on the filtered images $\tilde{\image}_\height$ and $\tilde{\image}_\distxy$, the surface normal vector can be approximated for each measurement element.
Here, the image representation has strong advantages over other representations such as unordered point sets as it implicitly defines a neighborhood.
\begin{figure}[t]
    \fontsize{8pt}{8pt}\selectfont
    \centering
    \captionsetup[subfigure]
    {skip=-59pt,slc=off,margin={1pt,0pt}}
    \subcaptionbox{\label{fig:n_ex_depth}}
    {\includesvg[width=0.24\linewidth]{normals_est_depth}}
    \subcaptionbox{\label{fig:n_ex_sphere}}
    {\includegraphics[width=0.24\linewidth]{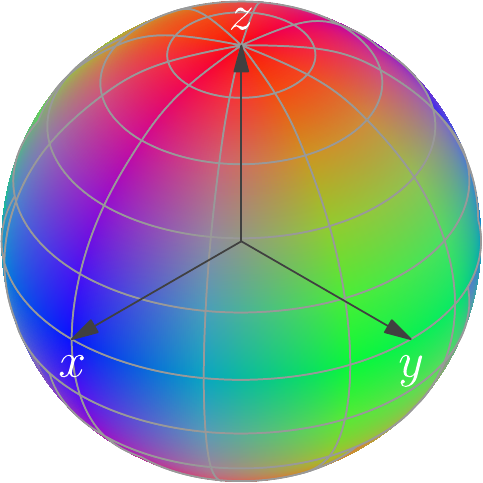}}
    \subcaptionbox{\label{fig:n_ex_naive}}
    {\includesvg[width=0.24\linewidth]{normals_est_br}}
    \subcaptionbox{\label{fig:n_ex_adapt}}
    {\includesvg[width=0.24\linewidth]{normals_est_adapt}}
    \captionsetup{subrefformat=parens}
    \caption{Pixel-wise surface normal vector calculation based on neighboring pixels: \subref{fig:n_ex_depth} an except of a range image with foreground (blue) and background (yellow-red), 
    \subref{fig:n_ex_sphere} the color map on the unit sphere used to visualize surface normal vectors,
    \subref{fig:n_ex_naive} naive normal vector calculation,
    \subref{fig:n_ex_adapt} normal vector calculation using nearest neighboring pixels.}
    \label{fig:neighbor_selection}
\end{figure}
The selection of neighboring pixels used for approximating the surface normal can be crucial.
When considering the same neighborhood in each measurement element, the considered values might represent surfaces of different entities and the surface normal calculation will be erroneous, see \Cref{fig:n_ex_naive}.
To minimize this effect, the considered neighborhood is adapted based on the range measurements.
To find the horizontally adjacent pixel $\cell_h$, the next pixel to the left and to the right are considered and the one that has the smaller Euclidean distance in the \ac{3D} \cartesian{} space to the measurement elements is chosen.
For the vertically adjacent pixel $\cell_v$, the next pixel below and above are considered.
In case no direct neighbors can be found, the second next pixels are considered and so forth until a maximal neighborhood size is reached.
This might be necessary in case no reliable measurements were recorded at neighboring locations.
The selection of adjacent pixels used for the normal vector calculation is demonstrated in \Cref{fig:n_ex_adapt}.

Given the filtered images $\tilde{\image}_\height$ and $\tilde{\image}_\distxy$, the sensor calibration $\calib$ and the pixels $\cell, \cell_h, \cell_v\in\grid_{uv}$, the \cartesian{} coordinates $\pos, \pos_h, \pos_v\in\real^3$ can be calculated and the normal vector $\nvec$ can be determined by computing and normalizing the corresponding cross product
\begin{equation}
    \nvec = \trans{(n_1, n_2, n_3)} = \frac{(\pos_h - \pos) \times (\pos_v - \pos)}{\norm{(\pos_h - \pos) \times (\pos_v - \pos)}}. \label{eq:cross}
\end{equation}
Based on the surface normal vector $\nvec$, an occupancy weight $w_\occ$ is calculated.
A logistic function centered around $\frac{\pi}{4}$ is applied to $\arccos(n_3)$, i.e. the angle between $\nvec$ and the North Pole $\trans{(0, 0, 1)}$ as
\begin{equation}
    w_\occ = \frac{1}{1 + \exp(-k(\arccos(n_3) - \frac{\pi}{4}))}, \label{eq:w_occ}
\end{equation}
where $k\in\real_{>0}$ is a scaling factor.
The logistic function is parametrized so that we have $w_\occ=0.5$ for $\arccos(n_3) = \frac{\pi}{4}$ which is based on the geometric consideration that a surface is considered blocking if its slope exceeds 45°.

The credibility of the calculated surface normal vector depends on the distance between the measurement element and its neighbors.
If the distance is small it might be dominated by inaccurate range measurements and the resulting normal vector might be disturbed.
Therefore, a normal vector confidence value
\begin{equation}
    \conf_{\nvec} = \frac{1}{1 + \exp(-k^{\prime}(\min(\|\pos_h - \pos\|, \|\pos_v - \pos\|) - \sigma_{\range}))} \label{eq:nvec_conf}
\end{equation}
is calculated quantifying if the minimal distance to the neighboring pixel coordinates is smaller or larger than the standard deviation $\sigma_{\range}$ of the range measurement.
Finally, the occupancy probability is calculated as
\begin{equation}
    p_\occ = \conf_{\nvec} \cdot w_\occ. \label{eq:pr_occ}
\end{equation}

Besides the image $\image_\occ$ containing the occupancy probabilities, the distance to ground is estimated for each measurement element.
This information is needed to exclude measurements outside the considered driving corridor and to clip measurement rays that intersect with the free space corridor boundaries.
One option is to explicitly estimate a ground surface using a parametric model such as a plane or a \ac{2D}  B-spline in the \cartesian{} space as in \cite{Wirges2021}.
However, estimating a parametric model often comes along with solving an optimization problem with computationally time-consuming numerical solvers.
Instead, we estimate the distance to ground directly for each measurement element.
Therefore, measurement elements that belong to measurement rays hitting the ground surface are classified, and the measured height is assigned directly.
Then the missing regions, i.e. rays that have been reflected by objects, have to be filled.
One best guess for this is to traverse the measurement image column-wise and propagate the last known ground height.
The height is propagated, if the corresponding measurement element was not reflected on the ground.
This classification of a measurement element is done based on the following heuristic considerations:

Let $\cell_{uv}\in\grid_{uv}$ be the current measurement element.
In case $\cell_{uv}$ is not located in the bottom row, let $\cell_{uv}^{-1}\in\grid_{uv}$ be the element located below $\cell_{uv}$.
Based on similar geometric considerations as in the calculations of the occupancy weight $w_\occ$ (\Cref{eq:w_occ}), the measurement element is classified as obstacle, if
\begin{itemize}    
    \item the angle between the surface normal vector $\nvec$ and the North Pole $\trans{(0, 0, 1)}$ exceeds 45°, i.e. $\arccos(n_3) > \frac{\pi}{4}$, or
    \item $\cell_{uv}$ is located in the bottom row and the vertical component of the detection coordinate minus the sensor height exceeds a given threshold, or
    \item $\cell_{uv}$ is not located in the bottom row and the Euclidean distance in the \ac{3D} \cartesian{} space to the measurement in $\cell_{uv}\in\grid_{uv}$ is smaller than the distance to the measurement in $\cell_{uv}^{-1}\in\grid_{uv}$.
\end{itemize}
After the first obstacle detection was found in a column, subsequent elements are only classified as ground, if the upper conditions are not fulfilled and the measured height has decreased compared to the measurement in $\cell_{uv}^{-1}$.
This is crucial to prevent classifying horizontal surfaces on obstacles as ground.

\subsubsection{\MG{}} \label{subsec:gm_measurement_grid}

On the \mg{} $\grid_{ur}$, measurement elements are assigned to the occupancy hypotheses $\omega\subseteq\frameo$ and ground hypotheses $\omega\subseteq\frameg$ and spatial uncertainty is models by applying the inverse sensor model.
The multi-layer grid map
\begin{equation}
    \hist_\measurement \colon \grid_{ur} \times \left(\powerset{\frameg}\cup\powerset{\frameo}\right) \rightarrow \real
\end{equation}
accumulates measurement elements for each hypothesis $\omega\subseteq\frameo$ and $\omega\subseteq\frameg$, respectively, in the corresponding grid layer $\hist_\measurement(\,\cdot\,, \omega)$.
Based on \Cref{eq:bba,eq:not_rel_obj_cell}, the logarithms of the probabilities $\pr{m\nrightarrow\omega,\cell}$ are accumulated as
\begin{equation}
    \hist_\measurement(\cell, \omega) = \sum\limits_{\cell_{uv}\in\grid_{uv}} \log(\pr{m\nrightarrow\omega,\cell})
\end{equation}
in grid cell $\cell\in\grid_{ur}$ for the hypotheses $\omega\in\powerset{\frameg}\cup\powerset{\frameo} \setminus \ofree$.

For the free space hypothesis $\ofree\subset\frameo$, the ray permeability $\perm$ defined in \Cref{eq:permeability} is estimated.
Therefore, we calculate 
\begin{equation}
    \hist_\measurement(\cell_{ur}, \ofree) = \perm_{\cell_{ur}}.
\end{equation}
For this purpose, a \ac{3D} ray casting is applied based on the sensor intrinsics.
The rationale behind this is that the portion of space covered by a measurement ray between the reflecting surface and the sensor origin provides free space evidence.
\begin{figure}[!t]
    \centering
    \fontsize{8pt}{8pt}\selectfont
    \includesvg[width=\linewidth]{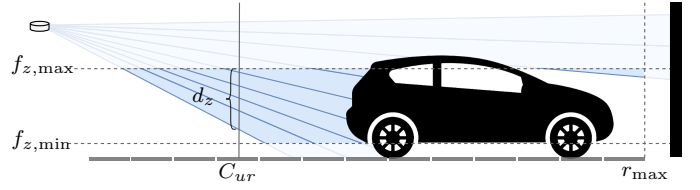}
    \caption{Calculation of the ray permeability $\perm_{\cell_{ur}}$ in a \mg{} cell $\cell_{ur}\in\grid_{ur}$. The rays are clipped according to the minimal height $d_{z, \min}$, the maximal height $d_{z, \max}$ and maximal range $r_{\max}$.}
    \label{fig:raycast}
\end{figure}
Each measurement ray contributes to the observed height where the contribution is defined by the ray divergence as sketched in \Cref{fig:raycast}.
For this purpose, we approximately model the vertical ray coverage to be dense, i.e. it is assumed that there are no gaps between the rays.
This is justified by the high vertical resolution of the sensors used in this work due to which only insignificant entities may be missed at full ray coverage.
Consequently, the ray permeability $\perm_{\cell_{ur}}$ in grid cell $\cell_{ur}\in\grid_{ur}$ is computed as
\begin{equation}
    \perm_{\cell_{ur}}  = \frac{d_z}{f_{z,\max} - f_{z, \min}},\quad d_z = \sum\limits_{\cell_{uv}\in\grid_{uv}} \func{z}(r_m),
\end{equation}
where $\func{z}$ is a function calculating the height of a measurement ray at range $r_m$.
Each measurement ray is clipped according to $f_{z,\min}$, $f_{z,\max}$ as only parts of the ray that traverse the defined free space corridor contribute to the free space estimation.
Positive and negative ray heights are mapped into the underlying grid cells to mark end and start points of the clipped measurement rays.
The missing gaps can then be filled by simply computing the running sum and correcting the ray heights to account for the ray divergence.

\subsubsection{\CG{}} \label{subsec:gm_cartesian_grid}

After calculating the \mg{} map $\hist_\measurement$, the second change of coordinates from the \mg{} $\grid_{ur}$ to the \cg{} $\gridxy$ is applied.
Therefore, cell values $\hist_\measurement(\cell, \omega)$ must be transformed properly into the \cartesian{} representation $\hist_{xy}(\cell, \omega)$.
For all hypotheses $\omega\neq\ofree$ except free space, this is done by integrating $\hist_\measurement$ over $\tf_{xy}^{ur}(\cell)\subset\rect_{ur}$ as
\begin{equation}
    \hist_{xy}(\cell, \omega) = \int\limits_{\tf_{xy}^{ur}(\cell)} \hist_\measurement(x, \omega)\,\dx{x}. \label{eq:ur_to_xy}
\end{equation}
For free space $\omega = \ofree$, the average over $\tf_{xy}^{ur}(\cell)$ is calculated as
\begin{equation}
    \hist_{xy}(\cell, \ofree) = \frac{1}{\leb{\tf_{xy}^{ur}(\cell)}}\int\limits_{\tf_{xy}^{ur}(\cell)} \hist_\measurement(x, \omega)\,\dx{x}. \label{eq:ur_to_xy_free}
\end{equation}
Here, $\leb{\,\cdot\,}$ denotes the \ac{2D}  Lebesgue-measure which calculates the area of $\tf_{xy}^{ur}(\cell)$.

The \cg{} map $\hist_{xy}$ is subsequently transformed to a consistent \ac{BBA} represented by the grid map $\gm_\measurement$.
Following \Cref{eq:bba,eq:not_rel_obj_cell,eq:bba_free}, it is computed as
\begin{equation}
    \gm_\measurement(\cell, \omega) = k\,(1 - \exp(\hist_{xy}(\cell, \omega))),
\end{equation}
for $\omega \subseteq \frameo\text{ or }\omega \in \frameg$, and 
\begin{equation}
    \gm_\measurement(\cell, \ofree) =  \biggl(1 - \sum\limits_{\psi\neq\ofree} \gm_\measurement(\cell, \psi) \biggr)\,\hist_{xy}(\cell, \omega)
\end{equation}
for the free space hypothesis $\ofree\subset\frameo$.
Here,
\begin{equation}
    k = \frac{1 - \exp\Bigl(\sum\limits_{\omega \subseteq \frameo} \hist_{\measurement}(\cell, \omega)\Bigr)}{\sum\limits_{\omega \subseteq \frameo} 1 - \exp\left(\hist_{\measurement}(\cell, \omega)\right)}
\end{equation}
is a normalization factor distributing \ac{BBA} masses equally to conflicting hypotheses.


\subsection{Grid Mapping Considering Sensor Modalities} \label{sec:sensor_gm_sensors}

Up to this point, the inverse sensor model $\pr{\cell \vert m}$ and the \mg{} $\grid{ur}$ were not specified as they depend on the sensor modalities.
In this section, we present the \mg{}s designed for \acp{lidar}, \glspl{monocam} and \glspl{stereocam} and two inverse sensor models.

\begin{figure}[!t]
    \centering
    \fontsize{8pt}{8pt}\selectfont
    \includesvg[width=\linewidth]{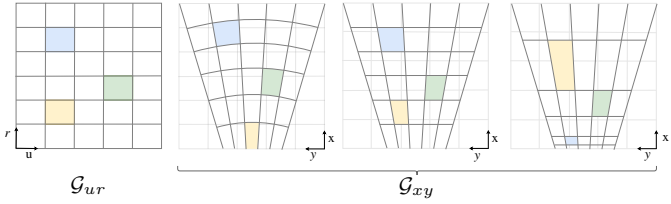}
    \caption{The \mg{} $\grid_{ur}$ on the left and the transformed \polar{}, u/distance and u/disparity grid $\tf_{ur}^{xy}(\grid_{ur})$. 
            Matching grid cell colors indicate the same grid cell in the source grid and the warped target grid.}
    \label{fig:m_grids}
\end{figure}
The \mg{} $\grid_{ur}$ used for \ac{lidar} is defined in \polar{} coordinates, i.e. the horizontal index $u$ corresponds to the angle location of the spinning laser and the range measurement is the measured distance projected to the xy-plane.
For measurements obtained from a \gls{monocam} where the distance was estimated directly in each pixel, an u/distance grid is used.
In case disparity estimates obtained from a \gls{stereocam} are mapped, an u/disparity grid is used.
\Cref{fig:m_grids} shows the relation between the \mg{} $\grid_{ur}$ and the \cg{} $\gridxy$ for all three \mg{}s used in this work.
The cell values in the \mg{} map $\hist_\measurement$ are transformed to \cartesian{} coordinates using \Cref{eq:ur_to_xy,eq:ur_to_xy_free}.

The inverse sensor model is calculated in the \mg{} $\grid{ur}$.
Let a \mg{} cell $\cell = I_u\times I_r\in\grid_{ur}$ be divided into its $u$ and $r$ components.
We model the measurement $m$ to be uniformly distributed in $I_u$.
Hence, the inverse sensor model simplifies to 
\begin{equation}
    \pr{\cell \vert m} = \pr{r \vert r_m} = \int\limits_{r\in I_r} \func{f}_{r_m}(r),
\end{equation}
where $\func{f}_{r_m}$ is the \ac{PDF} of the range $r$ given the range measurement $r_m$.
\begin{figure}[!t]
    \centering
    \providecommand{\figheight}{0.7in}%
    \providecommand{\figwidth}{1.25in}%
    \fontsize{7pt}{7pt}\selectfont
    \subcaptionbox{The \ac{PDF} for the Gaussian inverse sensor model.\label{fig:ism_gauss}}%
    [.48\linewidth]{
\begin{tikzpicture}

\definecolor{color0}{rgb}{0,0.56078431372549,0.835294117647059}

\begin{axis}[
axis line style={white},
height=\figheight,
minor xtick={},
minor ytick={},
tick align=outside,
tick pos=left,
width=\figwidth,
x grid style={white!79.6078431372549!black},
xlabel={\(\displaystyle r\)},
xmajorgrids,
xmin=-0.33, xmax=0.33,
xtick style={color=black},
xtick={-0.2,0,0.2},
xticklabels={
  \(\displaystyle r_m-0.2r\),
  \vphantom{0}\(\displaystyle r_m\),
  \(\displaystyle r_m+0.2r\)
},
y grid style={white!79.6078431372549!black},
ylabel={\(\displaystyle \mathrm{f}_{\mathcal{N}, r_m}(r)\)},
ymajorgrids,
ymin=0, ymax=4.5,
ytick style={color=black},
ytick={0,2,4,6}
]
\addplot [ultra thick, color0]
table {%
-0.3 0.04431848411938
-0.293939393939394 0.053057884252876
-0.287878787878788 0.0632877642858276
-0.281818181818182 0.0752132534870932
-0.275757575757576 0.0890581751341603
-0.26969696969697 0.105064985221502
-0.263636363636364 0.123494327915211
-0.257575757575758 0.144624147976342
-0.251515151515152 0.168748301866222
-0.245454545454545 0.196174612857474
-0.239393939393939 0.227222321450954
-0.233333333333333 0.262218890937095
-0.227272727272727 0.301496139168006
-0.221212121212121 0.345385681567305
-0.215151515151515 0.394213687040283
-0.209090909090909 0.448294967585927
-0.203030303030303 0.507926443758041
-0.196969696969697 0.573380051248129
-0.190909090909091 0.644895178209328
-0.184848484848485 0.722670747823327
-0.178787878787879 0.806857085231636
-0.172727272727273 0.897547731413242
-0.166666666666667 0.994771387927487
-0.160606060606061 1.0984841946507
-0.154545454545455 1.20856255671499
-0.148484848484848 1.32479674584754
-0.142424242424242 1.44688550433753
-0.136363636363636 1.57443187618844
-0.13030303030303 1.70694047909048
-0.124242424242424 1.84381641233661
-0.118181818181818 1.98436596963085
-0.112121212121212 2.12779929210819
-0.106060606060606 2.27323505631361
-0.1 2.41970724519143
-0.093939393939394 2.56617399843244
-0.0878787878787879 2.71152848320478
-0.0818181818181818 2.85461166898405
-0.0757575757575758 2.994226832711
-0.0696969696969697 3.1291555647792
-0.0636363636363637 3.2581749943769
-0.0575757575757576 3.38007590643616
-0.0515151515151515 3.49368138372654
-0.0454545454545455 3.59786557812623
-0.0393939393939394 3.69157219619697
-0.0333333333333333 3.77383227692993
-0.0272727272727273 3.84378084457259
-0.0212121212121212 3.90067203701287
-0.0151515151515151 3.94389234004919
-0.00909090909090909 3.97297159931742
-0.00303030303030305 3.98759153353742
0.00303030303030305 3.98759153353742
0.00909090909090909 3.97297159931742
0.0151515151515151 3.94389234004919
0.0212121212121212 3.90067203701287
0.0272727272727273 3.84378084457259
0.0333333333333333 3.77383227692993
0.0393939393939394 3.69157219619697
0.0454545454545455 3.59786557812623
0.0515151515151515 3.49368138372654
0.0575757575757576 3.38007590643616
0.0636363636363637 3.2581749943769
0.0696969696969697 3.1291555647792
0.0757575757575757 2.994226832711
0.0818181818181818 2.85461166898405
0.0878787878787879 2.71152848320478
0.0939393939393939 2.56617399843244
0.1 2.41970724519143
0.106060606060606 2.27323505631361
0.112121212121212 2.12779929210819
0.118181818181818 1.98436596963085
0.124242424242424 1.84381641233661
0.13030303030303 1.70694047909048
0.136363636363636 1.57443187618844
0.142424242424242 1.44688550433753
0.148484848484848 1.32479674584754
0.154545454545455 1.20856255671499
0.160606060606061 1.0984841946507
0.166666666666667 0.994771387927487
0.172727272727273 0.897547731413243
0.178787878787879 0.806857085231636
0.184848484848485 0.722670747823327
0.190909090909091 0.644895178209328
0.196969696969697 0.573380051248129
0.203030303030303 0.507926443758041
0.209090909090909 0.448294967585927
0.215151515151515 0.394213687040283
0.221212121212121 0.345385681567306
0.227272727272727 0.301496139168006
0.233333333333333 0.262218890937095
0.239393939393939 0.227222321450954
0.245454545454545 0.196174612857474
0.251515151515152 0.168748301866222
0.257575757575758 0.144624147976342
0.263636363636364 0.123494327915211
0.26969696969697 0.105064985221502
0.275757575757576 0.0890581751341604
0.281818181818182 0.0752132534870932
0.287878787878788 0.0632877642858276
0.293939393939394 0.0530578842528761
0.3 0.04431848411938
};
\end{axis}

\end{tikzpicture}}\hfill
    \subcaptionbox{The \ac{PDF} for the Interval inverse sensor model.\label{fig:ism_rect}}
    [.48\linewidth]{
\begin{tikzpicture}

\definecolor{color0}{rgb}{0,0.56078431372549,0.835294117647059}

\begin{axis}[
axis line style={white},
height=\figheight,
minor xtick={},
minor ytick={},
tick align=outside,
tick pos=left,
width=\figwidth,
x grid style={white!79.6078431372549!black},
xlabel={\(\displaystyle r\)},
xmajorgrids,
xmin=-0.33, xmax=0.33,
xtick style={color=black},
xtick={-0.2,0.2},
xticklabels={
  \vphantom{0}\(\displaystyle r_m\),
  \vphantom{0}\(\displaystyle r_{m^\prime}\)
},
y grid style={white!79.6078431372549!black},
ylabel={\(\displaystyle \mathrm{f}_{I, r_m}(r)\)},
ymajorgrids,
ymin=0, ymax=1.4,
ytick style={color=black},
ytick={0,1},
yticklabels={\(\displaystyle 0\),\(\displaystyle \frac{1}{\delta_r}\)}
]
\addplot [ultra thick, color0]
table {%
-0.3 0
-0.293939393939394 0
-0.287878787878788 0
-0.281818181818182 0
-0.275757575757576 0
-0.26969696969697 0
-0.263636363636364 0
-0.257575757575758 0
-0.251515151515152 0
-0.245454545454545 0
-0.239393939393939 0
-0.233333333333333 0
-0.227272727272727 0
-0.221212121212121 0
-0.215151515151515 0
-0.209090909090909 0
-0.203030303030303 0
-0.196969696969697 1
-0.190909090909091 1
-0.184848484848485 1
-0.178787878787879 1
-0.172727272727273 1
-0.166666666666667 1
-0.160606060606061 1
-0.154545454545455 1
-0.148484848484848 1
-0.142424242424242 1
-0.136363636363636 1
-0.13030303030303 1
-0.124242424242424 1
-0.118181818181818 1
-0.112121212121212 1
-0.106060606060606 1
-0.1 1
-0.093939393939394 1
-0.0878787878787879 1
-0.0818181818181818 1
-0.0757575757575758 1
-0.0696969696969697 1
-0.0636363636363637 1
-0.0575757575757576 1
-0.0515151515151515 1
-0.0454545454545455 1
-0.0393939393939394 1
-0.0333333333333333 1
-0.0272727272727273 1
-0.0212121212121212 1
-0.0151515151515151 1
-0.00909090909090909 1
-0.00303030303030305 1
0.00303030303030305 1
0.00909090909090909 1
0.0151515151515151 1
0.0212121212121212 1
0.0272727272727273 1
0.0333333333333333 1
0.0393939393939394 1
0.0454545454545455 1
0.0515151515151515 1
0.0575757575757576 1
0.0636363636363637 1
0.0696969696969697 1
0.0757575757575757 1
0.0818181818181818 1
0.0878787878787879 1
0.0939393939393939 1
0.1 1
0.106060606060606 1
0.112121212121212 1
0.118181818181818 1
0.124242424242424 1
0.13030303030303 1
0.136363636363636 1
0.142424242424242 1
0.148484848484848 1
0.154545454545455 1
0.160606060606061 1
0.166666666666667 1
0.172727272727273 1
0.178787878787879 1
0.184848484848485 1
0.190909090909091 1
0.196969696969697 1
0.203030303030303 0
0.209090909090909 0
0.215151515151515 0
0.221212121212121 0
0.227272727272727 0
0.233333333333333 0
0.239393939393939 0
0.245454545454545 0
0.251515151515152 0
0.257575757575758 0
0.263636363636364 0
0.26969696969697 0
0.275757575757576 0
0.281818181818182 0
0.287878787878788 0
0.293939393939394 0
0.3 0
};
\end{axis}

\end{tikzpicture}}
    \caption{The \acp{PDF} of the range $r$ used in the two presented inverse sensor models.}
    \label{fig:ism}
\end{figure}
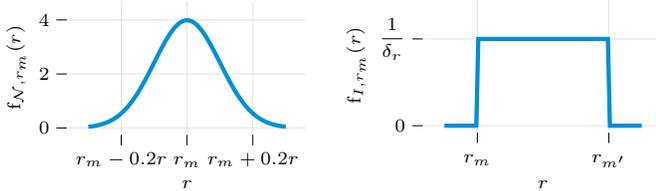
In this work, we present two options for $\func{f}_{r_m}$:
\paragraph{The Gaussian Model} 
We model the range to be normally distributed with mean $r_m$ and standard deviation $\sigma_r$, i.e.
\begin{equation}
    \func{f}_{\mathcal{N}, r_m}(r) = \frac{1}{\sqrt{2\pi}\,\sigma_r}\,\exp\left(-\frac{1}{2}\left(\frac{r-r_m}{\sigma_r} \right)^2\right).
\end{equation}
In the remainder of this thesis, we refer to this model as Gaussian inverse sensor model $\fpr_{\mathcal{N}}(\cell\vert m)$.
It is sketched in \Cref{fig:ism_gauss} with standard deviation $\sigma_r=0.1r$.
\paragraph{The Interval Model} 
The measurement element $m$ covers the whole interval $[r_m, r_{m^\prime}]$, i.e.
\begin{equation}
    \func{f}_{I, r_m}(r) =  \begin{cases}
        \frac{1}{\delta_r}, & \text{ if }  r \in [r_m, r_{m^\prime}] \\
        0, & \text{ else,}
    \end{cases}
\end{equation}
where $\delta_r$ is the length of the range interval $I_r$ of one grid cell $\cell\in\grid_{ur}$ and $r_{m^\prime}$ is the range measurement in the \sg{} cell $\cell^\prime = (u, v+1)$ that is vertically adjacent to the considered measurement element $m$ in $\cell$.
Note that $\func{f}_{I, r_m}(r)$ is normalized so that it integrates to one over the range interval $I_r$ of a fully supported grid cell.
This model is based on the assumption that the measurement $\measurement$ partitions the measured surface which means that there are no gaps between areas on the world surface covered by rows in the \sg{} $\grid_{uv}$.
In the remainder of this thesis, we refer to this model as Interval inverse sensor model $\fpr_{I}(\cell\vert m)$.
It is sketched in \Cref{fig:ism_rect}.
We apply $\func{f}_{I, r_m}$ only in the calculation of the \ac{BBA} on the ground hypotheses $\frameg$ as the model assumption is violated for occupying surfaces.
Furthermore, we propose using it only for \gls{cam} measurements and not for \ac{lidar} measurements as \ac{lidar} scan lines are usually not adjacent.
In comparison, the gaps between pixel rows in camera sensor used in automotive applications are negligible.

\section{Experiments} \label{sec:experiments}

Our proposed grid mapping framework is validated using the Kitti Vision Benchmark \cite{Geiger2012CVPR} and the semantic \ac{lidar} point cloud labels from the \gls{semantickitti} dataset extension \cite{behley2019iccv}.
They contain measurements from a \lidarkitti{} \ac{lidar} scanner with surround view, one RGB and one grayscale \gls{stereocam} setup pointing to the front, the sensor calibrations and 6D ego pose annotations.
In this work, the measurements from the \lidarkitti{} and the stereo RGB camera setup are processed with the presented grid mapping pipeline.

\subsection{Qualitative Results} \label{subsec:gm_eval_qualitative}

In the \sg{}, surface normal vectors are calculated based on the range estimates and occupancy probabilities are deduced as described in \Cref{subsec:gm_sensor_grid}.

\begin{figure}[!t]
    \fontsize{5pt}{5pt}\selectfont
    \begin{minipage}{\linewidth}
        \centering
        \includegraphics[width=\linewidth, height=0.08\linewidth]{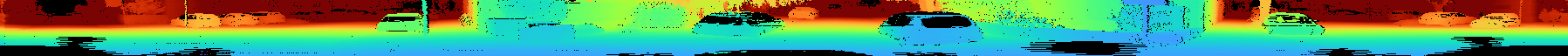}\\
        \colorbar{min=0,max=20,width=0.7}
    \end{minipage}\\[0.5em]
    \begin{minipage}{\linewidth}
        \centering
        \includegraphics[width=\linewidth, height=0.08\linewidth]{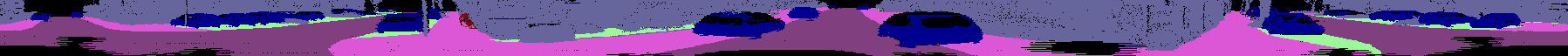}\\
        \colorbarpx
    \end{minipage}\\[0.8em]
    \begin{minipage}{\linewidth}
        \centering
        \includegraphics[width=\linewidth, height=0.08\linewidth]{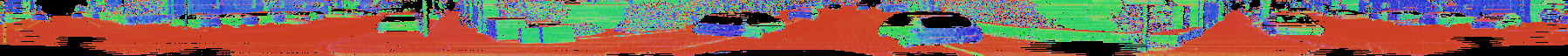}\\
        \colorbarnormals
    \end{minipage}\\[0.8em]
    \begin{minipage}{\linewidth}
        \centering
        \includegraphics[width=\linewidth, height=0.08\linewidth]{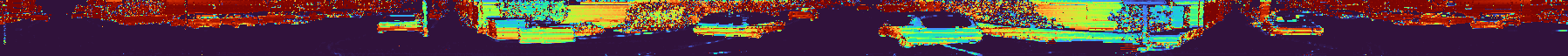}\\
        \colorbar{min=0,max=1,width=0.7}
    \end{minipage}
    \caption{Impressions of the \sg{} processing chain for a \ac{lidar} sensor reading. From top to bottom: Range image $\image_\range$, semantic segmentation $\image_\sem$, surface normal vectors $\image_\normals$ and occupancy probabilities $\image_\occ$}
    \label{fig:lidar_image_proc}
\end{figure}
\Cref{fig:lidar_image_proc} shows the image processing steps for a range image of a \lidarkitti{} measurement taken from the Kitti dataset.
The gaps in the range image result from shadows caused by sensors and antennas on top of the test vehicle or missing reflections due to surfaces with low reflectivity.
The latter tends to occur on surfaces at high distances larger than 100\si{\metre} or on dark surfaces such as black cars or windows.
The semantically annotated image in the second row contains the labels assigned to each \ac{lidar} detection in the \gls{semantickitti} dataset extension.
In the bottom row, the image containing the surface normal vectors colorcoded according to \Cref{fig:n_ex_sphere} and the resulting image $\image_\occ$ containing the occupancy probabilities are shown.
Recall that the occupancy probabilities in $\image_\occ$ scale with the orientation of the reflecting surface and thus do not segment the environment in objects and ground.
Consequently, there are areas on objects with low occupancy probability on horizontal surfaces such as the hood or the rooftop of cars.

\begin{figure}[!t]
    \centering
    \fontsize{5pt}{5pt}\selectfont
    \begin{minipage}[t]{.48\linewidth}
        \centering
        \includegraphics[width=\linewidth]{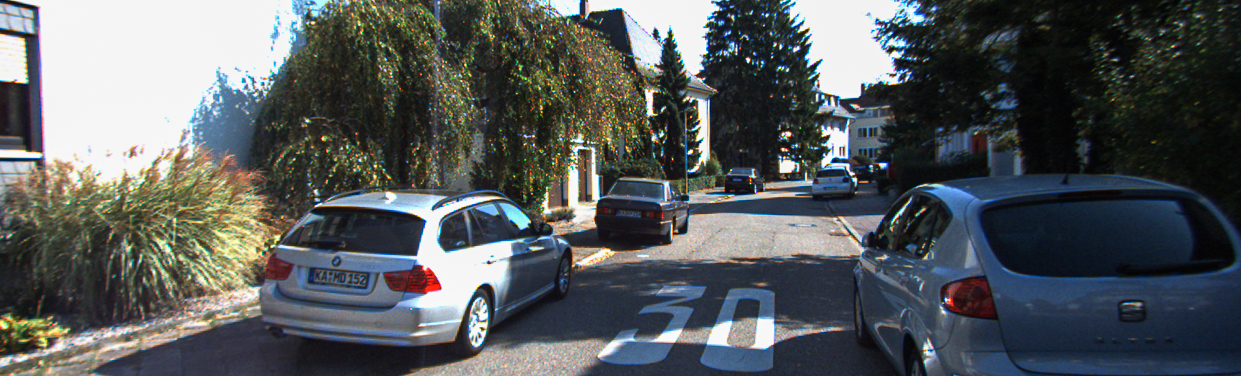}
    \end{minipage}\hfill
    \begin{minipage}[t]{.48\linewidth}
        \centering
        \includegraphics[width=\linewidth]{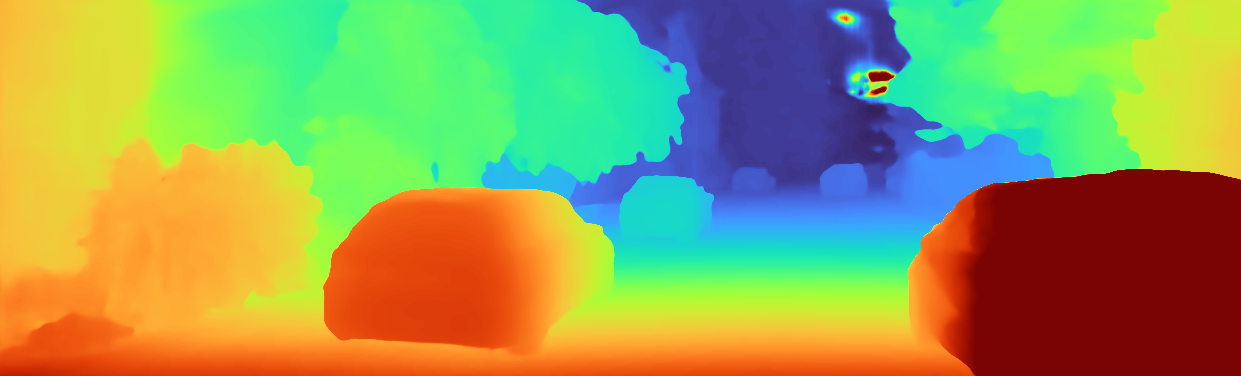}
        \colorbar{min=2,max=80,width=0.85}
    \end{minipage} \\[-0.6em]
    \begin{minipage}[t]{.25\linewidth}
        \centering
        \colorbarpxslim
    \end{minipage}
    \begin{minipage}[t]{.48\linewidth}
        \centering
        \includegraphics[width=\linewidth]{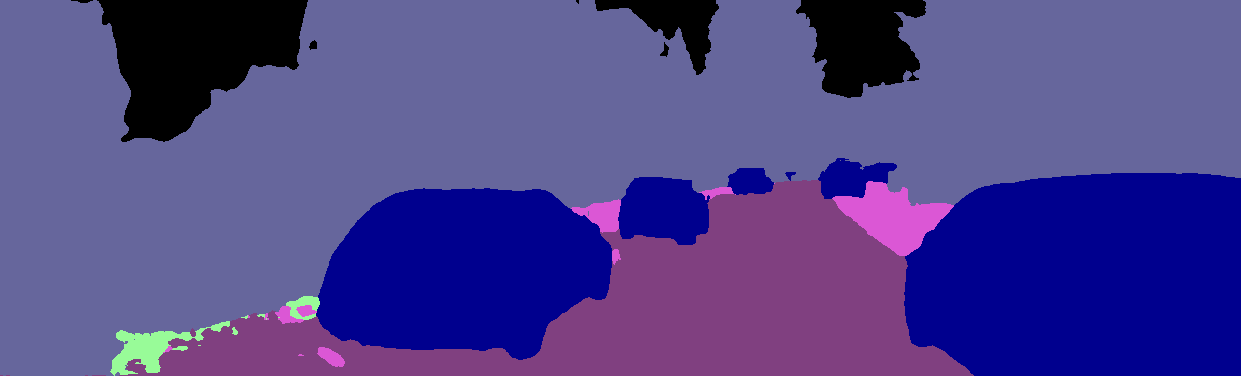}
    \end{minipage} \\[0.4em]
    \begin{minipage}[t]{.48\linewidth}
        \centering
        \includegraphics[width=\linewidth]{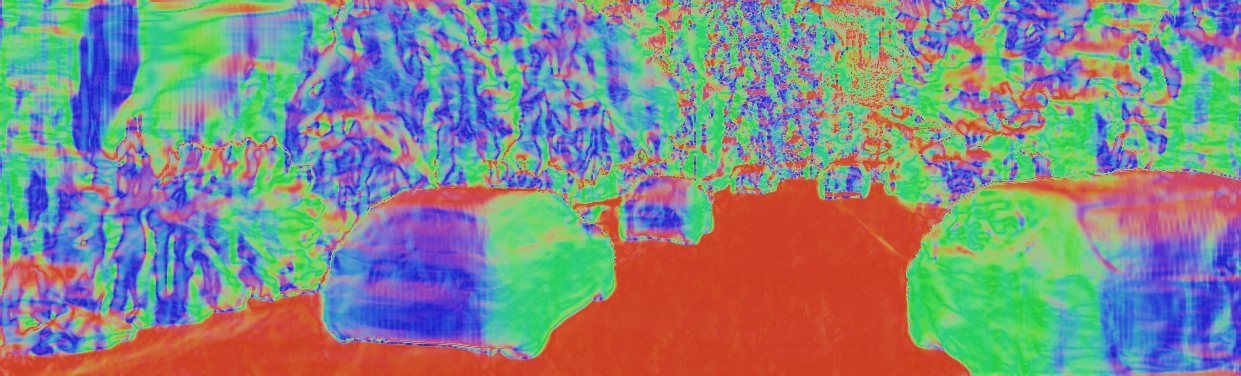}
        \colorbarnormals
    \end{minipage}\hfill
    \begin{minipage}[t]{.48\linewidth}
        \centering
        \includegraphics[width=\linewidth]{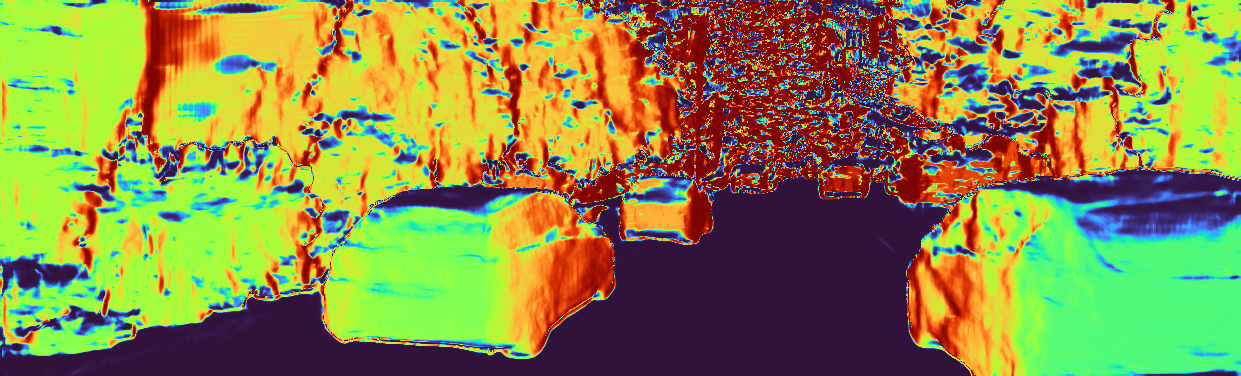}
        \colorbar{min=0,max=1,width=0.9}
    \end{minipage}
    \caption{Impressions of the \sg{} processing chain for a \gls{stereocam} sensor reading. From left to right, top to bottom: RGB image, disparity image $\image_\range$, semantic segmentation $\image_\sem$, surface normal vectors $\image_\normals$ and occupancy probabilities $\image_\occ$}
    \label{fig:stereo}
\end{figure}
\Cref{fig:stereo} shows the image processing results for an image pair of the \gls{stereocam} in the Kitti Tracking dataset.
The image view, shows the center part of the \ac{lidar} range image shown in \Cref{fig:lidar_image_proc}.
The disparity image depicted on the top right was estimated using the guided aggregation net for stereo matching presented by \citet{ganet}.
This is a well performing stereo disparity estimator generating dense disparity maps that comes along with a real-time capable implementation running at 15-20 \ac{FPS}.
The pixelwise semantically labelled image in the middle row was obtained by feeding the RGB image recorded by the left camera into the network presented by \citet{zhu2019improving}.
The inference on this network is not real-time capable, but similarly performing, real-time capable alternatives haven been proposed e.g. in \cite{hong2021deep}.
In the bottom row, the surface normal vectors $\image_\normals$ and the corresponding occupancy probability image $\image_\occ$ are shown.

\begin{figure}[!t]
    \fontsize{5pt}{5pt}\selectfont
    \begin{subfigure}[b]{\linewidth}
        \centering
        \includegraphics[width=0.5\linewidth]{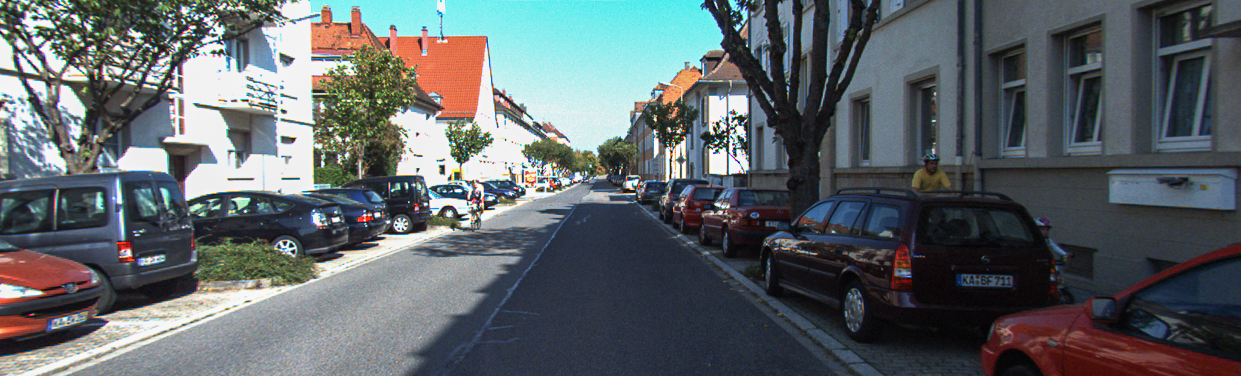}
        \caption{Image taken by the front left color camera.}
        \label{fig:lidar_image}
    \end{subfigure} \\[0.5em]
    \colorbarevslim
    \subcaptionbox{Resulting grid map without semantic estimates.\label{fig:bba_vis}}%
    [\linewidth]{
        \centering
        \widegrid{file=lidar/bba_vis_kitti_odom_00_3618,width=0.54}
        \hspace{-0.6em}
        \widegridcam{file=stereo/bba_vis_kitti_odom_00_3618,width=0.324}\vspace{-0.3em}
    } \\[0.7em]
    \subcaptionbox{Resulting grid map with semantic estimates using the inverse sensor model $\fpr_{\mathcal{N}}(\cell\vert m)$ for both ground and occupancy.\label{fig:bba_vis_sem}}%
    [\linewidth]{
        \centering
        \widegrid{file=lidar/bba_vis_sem_kitti_odom_00_3618,width=0.54}
        \hspace{-0.5em}
        \widegridcam{file=stereo/bba_vis_sem_kitti_odom_00_3618,width=0.324}\vspace{-0.3em}
    } \\[0.7em]
    \subcaptionbox{Resulting grid map with semantic estimates using the interval sensor model $\fpr_{I}(\cell\vert m)$ for occupancy and $\fpr_{I}(\cell\vert m)$ for the ground.\label{fig:bba_vis_sem_rect}}%
    [\linewidth]{
        \centering
        \widegrid{file=lidar/bba_vis_sem_rect_kitti_odom_00_3618,width=0.54}
        \hspace{-0.5em}
        \widegridcam{file=stereo/bba_vis_sem_rect_kitti_odom_00_3618,width=0.324}\vspace{-0.3em}
    }
    \caption{Resulting \ac{BBA} visualizations in the \cg{} $\gridxy$ using \ac{lidar} (left) and \gls{stereocam} measurements (right) with and without semantic estimates.}
    \label{fig:sensor_bba_vis}
\end{figure}
\Cref{fig:sensor_bba_vis} visualizes the resulting grid map estimated with the same measurements as in \Cref{fig:lidar_image_proc,fig:stereo}.
First, no semantic estimates are included, i.e. only the 360° \ac{lidar} scan and the stereo disparity map are processed.
\Cref{fig:bba_vis} show the occupancy probability after applying the pignistic transformation
\begin{equation}
    \pr{A} = \sum\limits_{B\subseteq\Omega}\frac{\vert A\cap B\vert}{\vert B\vert}\bba{B}.
    \label{eq:pignistic}
\end{equation}
Here, the gray values transition from white for zero to black denoting an occupancy probability of one.
In \Cref{fig:bba_vis_sem,fig:bba_vis_sem_rect} the resulting evidential grid map is visualized when additionally processing the semantic estimates from the pixelwise annotated images.
The applied color map shows both ground and object hypotheses and is a combination of the occupancy probability gray value map and a color coding of the semantic classes.
The color map was generated based on the following rules:
If both a ground hypothesis and an object hypotheses have been assigned a high \ac{BBA}, the object \ac{BBA} is visualized.
The color saturation scales with the assigned \ac{BBA}.
Above ground, a lower brightness indicates a low \ac{BBA} for \emph{free}.

The results clearly show the differences in modelling ground detections with the Gaussian inverse sensor model $\fpr_{\mathcal{N}}(\cell\vert m)$ in \Cref{fig:bba_vis_sem} and with the interval inverse sensor model $\fpr_{I}(\cell\vert m)$ in \Cref{fig:bba_vis_sem_rect}.
Whereas the \ac{BBA} estimation for the hypothesis street is sparse using the Gaussian model, evidence for the whole area of the street covered by the \sg{} is obtained when using the interval model.

\subsection{Occupancy estimation} \label{sec:gm_eval_occ}

One of the key tasks in the evidential grid map estimation is the deduction of occupancy evidence based on the measurements.
We demonstrate the differences when applying the presented framework to point sets and images.
Recall that two ways of defining the term occupied in a geometric manner were presented where \Cref{def:occupancy_1} is used for point sets and \Cref{def:occupancy_2} for images.
To make the evaluation representable, a subsequence of the Kitti odometry benchmark is chosen that was recorded on challenging terrain with altering height.
The \ac{BBA} for the hypothesis \emph{occupied} is calculated with one of the following three methods:
\begin{itemize}
    \item \emph{Flat world model.} Derive evidence for occupancy according to \Cref{def:occupancy_1} as described in \Cref{sec:sensor_gm_unstruct}. 
        The ground surface is modeled as a xy-plane $\{(x,y,z)\,\vert\,z=0\}$ in vehicle coordinates and the tolerance margin is set to $\delta_G = 0.3$\si{\metre}.
        The resulting \ac{BBA} is denoted as $\fbba_\flatw$.
    \item \emph{B-spline model.} Derive evidence for occupancy according to \Cref{def:occupancy_1} as described in \Cref{sec:sensor_gm_unstruct}. 
        The ground surface is represented by the uniform B-spline model proposed by \citet{Wirges2021} and the tolerance margin is set to $\delta_G = 0.3$\si{\metre}.
        The resulting \ac{BBA} is denoted as $\fbba_\spline$.
    \item \emph{Surface normals.} Derive evidence for occupancy according to \Cref{def:occupancy_2} as described in \Cref{sec:sensor_gm_struct}.
        The resulting \ac{BBA} is denoted as $\fbba_\normals$.
\end{itemize}
To evaluate the resulting \ac{BBA} for a cell being occupied the following \acp{BBA} are calculated:
\begin{itemize}
    \item The reference \ac{BBA} $\fbba_{\mathrm{ref}}$ calculated using \Cref{eq:bba} where the occupancy probability is set to 
    \begin{numcases}{p_{\occ} =}
        1, & if $\omega_{\mathrm{ref}} \subseteq \ooccupied$, \nonumber \\
        0, & else, \nonumber
    \end{numcases}
    where $\omega_{\mathrm{ref}}$ is the semantic label added in the \gls{semantickitti} dataset extension.
    \item The \ac{BBA} $\fbba_{\mathrm{all}}$ containing all detections, i.e. the occupancy weight is $p_{\occ} = 1$.
    \item The three estimated \acp{BBA} $\fbba_\flatw$, $\fbba_\spline$ and $\fbba_\normals$ calculated as described above.
\end{itemize}
Note that the reference \ac{BBA} $\fbba_{\mathrm{ref}}$ is not a ground truth classification based on the same geometric cues as used in the estimation.
As opposed to defining occupancy based on geometric constraints as in \Cref{def:occupancy_1,def:occupancy_2}, the reference \ac{BBA} deduces occupancy based on semantic constraints.
In order to create a ground truth \ac{BBA} based on geometric constraints, a complete \ac{3D} surface model of the environment would be required.
However, the comparison considered here still yields interpretable information on the performance of the \ac{BBA} estimation.
Based on $\fbba_{\mathrm{ref}}$, $\fbba_{\mathrm{all}}$ and $\fbba_i$, $i\in\{\flatw, \spline, \normals\}$, the confusion metrics
\begin{align*}
    \xi_{\mathrm{TP},i} &= \tilde{\fbba}_i(\ooccupied)\,\tilde{\fbba}_{\mathrm{ref}}(\ooccupied)\,\fbba_{\mathrm{all}}(\ooccupied), \\
    \xi_{\mathrm{FP},i} &= \tilde{\fbba}_i(\ooccupied)\,(1 - \tilde{\fbba}_{\mathrm{ref}}(\ooccupied))\,\fbba_{\mathrm{all}}(\ooccupied), \\
    \xi_{\mathrm{FN},i} &= (1 - \tilde{\fbba}_i(\ooccupied))\,\tilde{\fbba}_{\mathrm{ref}}(\ooccupied)\,\fbba_{\mathrm{all}}(\ooccupied), \\
    \xi_{\mathrm{TN},i} &= (1 - \tilde{\fbba}_i(\ooccupied))\,(1 - \tilde{\fbba}_{\mathrm{ref}}(\ooccupied))\,\fbba_{\mathrm{all}}(\ooccupied)
\end{align*}
are defined per grid cell $\cell\in\gridxy$, where 
\begin{equation}
    \tilde{\fbba}_i(\ooccupied) = \frac{\fbba_i(\ooccupied)}{\fbba_{\mathrm{all}}(\ooccupied)},\quad x\in\{\flatw, \spline, \normals, \mathrm{ref}\}
\end{equation}
is the part of $\fbba_{\mathrm{all}}(\ooccupied)$ that was classified as occupied.
The confusion rates on the whole grid $\gridxy$ are then calculated as
\begin{equation}
    \xi_{j,i} = \frac{\sum\limits_{\cell\in\gridxy} \xi_{j,i}(\cell)}{\sum\limits_{j\in J}\sum\limits_{\cell\in\gridxy} \xi_{\mathrm{TP},i}(\cell) },
\end{equation}
where $j\in J = \{\mathrm{TP},\mathrm{FP},\mathrm{FN},\mathrm{TN}\}$.
A high false positive rate $\xi_{\mathrm{FP},i}$ indicates that measurement elements $m$ with attached semantic label $\omega_{\mathrm{ref}}\in\frameg$ contributed to a high \ac{BBA} $\bba{\ooccupied}$ whereas a high false negative rate $\xi_{\mathrm{FN},i}$ indicates that measurement elements with semantic label $\omega_{\mathrm{ref}}\in\ooccupied$ had little contribution to $\bba{\ooccupied}$.

\begin{figure*}[!t]
    \centering
    \captionsetup{subrefformat=parens}
    \fontsize{8pt}{8pt}\selectfont
    \def\figa{\begin{minipage}[t]{0.33\linewidth}\hfill\includegraphics[width=0.92\linewidth]{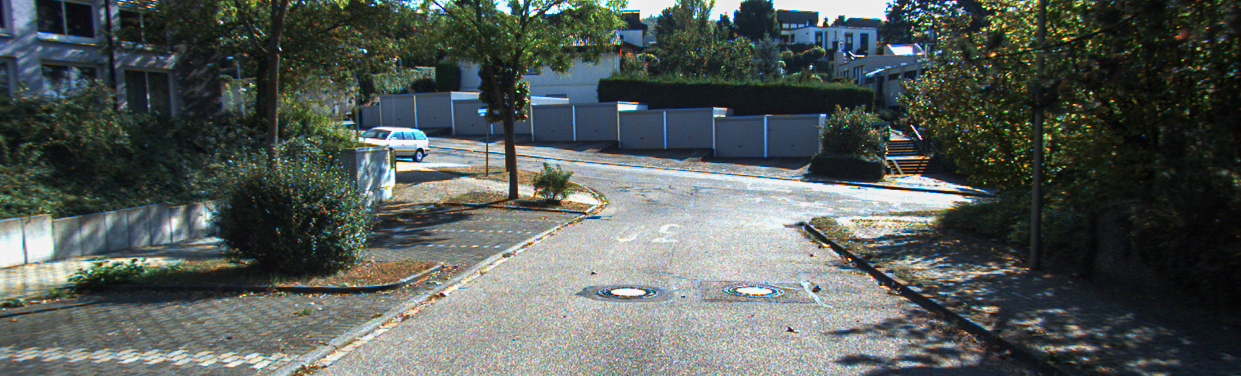}\end{minipage}}
    \def\figb{\begin{minipage}[t]{0.329\linewidth}\centering\fontsize{5pt}{5pt}\selectfont\widegrid{file=figures/lidar/ground_height_02528,width=0.92}\\\colorbar{min=-1,max=2.5,width=0.7}\end{minipage}}
    \def\figc{\begin{minipage}[t]{0.66\linewidth}\centering\providecommand{\figheight}{1.5in}\providecommand{\figwidth}{0.8\linewidth}\input{figures/lidar/occ_weights.tex}\end{minipage}}
    \def\capa{Image taken by the front left color camera.}
    \def\capb{The maximal detected height of all \ac{lidar} detections located in a specific grid cell.}
    \def\capc{Confusion rates $\xi_{j,i}$ for the occupancy \acp{BBA} $\fbba_\flatw$, $\fbba_\spline$ and $\fbba_\normals$.}
    \savestack{\capfiga}{\subcaptionbox{\capa\label{fig:lidar_gound_image}}{\figa}}
    \savestack{\capfigb}{\subcaptionbox{\capb\label{fig:lidar_gound_height}}{\figb}}
    \savestack{\capfigc}{\subcaptionbox{\capc\label{fig:lidar_weights_conf_chart}}{\begin{minipage}[t]{0.66\linewidth}\centering\providecommand{\figheight}{1.5in}\providecommand{\figwidth}{0.8\linewidth}\input{figures/lidar/occ_weights.tex}\end{minipage}}}
    \def\hgap{0ex}
    \stackon%
      [\heightof{\begin{minipage}[t]{0.66\linewidth}\centering\providecommand{\figheight}{1.5in}\providecommand{\figwidth}{0.8\linewidth}\input{figures/lidar/occ_weights.tex}\end{minipage}}-\heightof{\figb}-16pt-\heightof{\capfiga}-\depthof{\capfiga}]%
      {\capfigb}{\capfiga}\hfill\hspace{\hgap}\capfigc%
    \fontsize{5pt}{5pt}\selectfont
    \colorbarblockingconftv\\
    \subcaptionbox{Flat world assumption.\label{fig:lidar_gound_flat}}%
    [0.329\linewidth]{
        \centering
        \widegrid{file=figures/lidar/ground_flat_02528,width=0.93}\vspace{-1em}
    }\hfill
    \subcaptionbox{Ground surface estimation with \cite{Wirges2021}.\label{fig:lidar_ground_spline}}%
    [0.329\linewidth]{
        \centering
        \widegrid{file=figures/lidar/ground_spline_02528,width=0.93}\vspace{-1em}
    }\hfill
    \subcaptionbox{Proposed method.\label{fig:lidar_ground_proposed}}%
    [0.329\linewidth]{
        \centering
        \widegrid{file=figures/lidar/ground_normals_02528,width=0.93}\vspace{-1em}
    }
    \caption{Evaluation of the occupancy evaluation on challenging, hilly terrain, see \subref{fig:lidar_gound_image} and \subref{fig:lidar_gound_height}. 
            The frame shown in \subref{fig:lidar_gound_image}, \subref{fig:lidar_gound_height}, \subref{fig:lidar_gound_flat}, \subref{fig:lidar_ground_spline} and \subref{fig:lidar_ground_proposed} is marked by the gray vertical line in \subref{fig:lidar_weights_conf_chart}.
            The comparison in the bottom row demonstrates that our proposed model minimizes false positive occurrences.}
    \label{fig:lidar_weights_conf}
\end{figure*}
\Cref{fig:lidar_weights_conf} shows the results when processing \ac{lidar} measurements from a Kitti sequence on challenging, hilly terrain.
\Cref{fig:lidar_gound_image} shows a specific frame in that sequence.
The ego vehicle enters a street with a significant incline that leads to a crossing on rather flat terrain.
The camera image of the front left camera shows that there is a significant change in the gradient of the ground surface in the vicinity of the ego vehicle.
This is further highlighted in \Cref{fig:lidar_gound_height} that shows the maximal detected height of all \ac{lidar} detections located in a grid cell.
Here, the \ac{lidar} detections were transformed into the vehicle coordinate system so that zero corresponds to the height below the ego vehicle.

\Cref{fig:lidar_weights_conf_chart} shows the confusion metrics $\xi_{j,i}$ for 200 consecutive frames in the Kitti sequence.
The frame shown in \Cref{fig:lidar_gound_image,fig:lidar_gound_height} is marked by the gray vertical line in \Cref{fig:lidar_weights_conf_chart}.
The height of the road surface with respect to a global reference coordinate system is visualized in gray behind the plots for the six metrics.
Here, all grid cells with assigned ground label \emph{other ground}, i.e. everything but street and sidewalk are excluded as occupancy derived from semantic properties might differ significantly from the geometric occupancy on other ground like meadows and other vegetation.
Including those areas would distort the evaluation results.
It can be seen that the false positive rate $\xi_{\mathrm{FP},\flatw}$ of the flat world model is heavily influenced by uneven terrain violating the flat world assumption.
In most of the frames, the false positive rate $\xi_{\mathrm{FP},\spline}$ is reduced to approximately zero for the B-spline model.
However, between frame 2510 and 2540, a significant rise of $\xi_{\mathrm{FP},\spline}$ can be observed.
The false positive rate $\xi_{\mathrm{FP},\normals}$ for the proposed surface normal vector-based method, on the other hand, stays almost constant close to zero in the whole test sequence.
As in the proposed method, occupancy evidence is not deduced for horizontal surfaces on objects such as car roofs, the false negative rate $\xi_{\mathrm{FN},\normals}$ is the highest almost throughout the whole sequence.
The false negative rate $\xi_{\mathrm{FN},\spline}$ indicates that the largest number of detections reflected on object surfaces were not missed in the B-spline model.
However, it should be emphasized that this is due to the differences between the two underlying occupancy concepts in \Cref{def:occupancy_1} and \Cref{def:occupancy_2}.

\Cref{fig:lidar_gound_flat} depicts the mapping result when applying grid mapping with point sets using the flat world model.
In the area around the junction there are many ground detections that are classified as obstacles contributing to a high occupancy mass in those grid cells.
On challenging terrain as in this scenario this is expected as the ground model cannot capture the actual ground geometry.
\Cref{fig:lidar_ground_spline} shows the results when applying grid mapping with point sets using the B-spline model.
The fact that the uniform B-spline is able to capture the real surface significantly better leads to fewer grid cells falsely classified as occupied.
However, there are still some grid cells on the ground that are classified as occupied.
This can be resolved when applying the proposed surface normal vector model.
The results are shown in \Cref{fig:lidar_ground_proposed} where a high \ac{BBA} for the hypothesis occupied is mostly obtained in areas where obstacles are assumed to be present.
One exception are curb stones.
As opposed to the other two models, the surface normals model classifies grid cells located at curb stones as \emph{occupied}.
They are visualized in blue in \Cref{fig:lidar_ground_proposed} as curb stones are labeled as sidewalk in the \gls{semantickitti} labels and thus no occupancy evidence is deduced in the calculation of the reference \ac{BBA} $\fbba_{\mathrm{ref}}$.
The results shown in \Cref{fig:lidar_ground_proposed} further show that missing detections in $\fbba_\normals$ are in fact almost entirely located within objects and thus are negligible in top-view object shape estimation.

\section{Conclusion}

We proposed a novel grid mapping pipeline for \ac{lidar} and \gls{cam} measurements containing two advancements compared to past publications.
First, we base our evidential grid maps on a dual evidential representation modeling semantic occupancy and ground separately.
Second, our framework utilizes a surface orientation-based occupancy evidence deduction making an additional ground surface estimation obsolete.
We demonstrated in challenging, hilly traffic scenarios that our method provides detailed and robust mapping results.
Our future work will focus on improving sensor data fusion results and combining measurements over time in a recursive estimator.
We think that especially the latter will benefit from the dual representation presented in this work.

\printbibliography%

\end{document}